\documentclass[sensors,article,submit,moreauthors,pdftex, letterpaper]{Definitions/mdpi} 

\firstpage{1} 
\pubvolume{1}
\issuenum{1}
\articlenumber{0}
\pubyear{2021}
\copyrightyear{2020}
\datereceived{} 
\dateaccepted{} 
\datepublished{} 

\usepackage{times}
\usepackage{epsfig}
\usepackage{amssymb}

\Title{A Deep Learning Framework for Recognizing both Static and Dynamic Gestures}

\TitleCitation{A Deep Learning Framework for Recognizing both Static and Dynamic Gestures}

\Author{Osama~Mazhar $^{1, 2, *}$ \orcidA{}, Sofiane~Ramdani $^{1}$ and~Andrea~Cherubini $^{1}$}

\AuthorNames{Osama Mazhar, Sofiane Ramdani and Andrea Cherubini}

\AuthorCitation{Mazhar, O.; Ramdani, S.; Cherubini, A.}

\address{%
$^{1}$ \quad LIRMM, Université de Montpellier, CNRS
		Montpellier, France. \\
$^{2}$ \quad Technische Universiteit Delft, Netherlands.}

\corres{Correspondence: o.mazhar@tudelft.nl}

\abstract{Intuitive user interfaces are indispensable to interact with the human centric smart environments. In this paper, we propose a unified framework that recognizes both static and dynamic gestures, using simple RGB vision (without depth sensing). This feature makes it suitable for inexpensive human-robot interaction in social or industrial settings.
We employ a pose-driven spatial attention strategy, which guides our proposed Static and Dynamic gestures Network \textendash~\textit{StaDNet}.
From the image of the human upper body, we estimate his/her depth, along with the region-of-interest around his/her hands.
The Convolutional Neural Network in \textit{StaDNet} is fine-tuned on a background-substituted hand gestures dataset.
It is utilized to detect 10 static gestures for each hand as well as to obtain the hand image-embeddings.
These are subsequently fused with the augmented pose vector and then passed to the stacked Long Short-Term Memory blocks.
Thus, human-centred frame-wise information from the augmented pose vector and from the left/right hands image-embeddings are aggregated in time to predict the dynamic gestures of the performing person.
In a number of experiments, we show that the proposed approach surpasses the state-of-the-art results on the large-scale \textit{Chalearn 2016} dataset. Moreover, we transfer the knowledge learned through the proposed methodology to the \textit{Praxis gestures} dataset, and the obtained results also outscore the state-of-the-art on this dataset.}

\begin{document}
\section{Introduction}
Modern manufacturing industry requires human-centered smart frameworks, which aim to focus on human abilities and not conversely demand humans to adjust to whatever technology.
In this context, gesture-driven user-interfaces tend to exploit human's prior knowledge and are vital for intuitive interaction of humans with smart devices \cite{li2018dynamic}.
Gesture recognition is a problem that has been widely studied for developing human-computer/machine interfaces with an input device alternative to the traditional ones (\textit{e.g.}, mouse, keyboard, teach pendants and touch interfaces).
Its applications
include robot control \cite{kofman2005teleoperation, tolgyessy2017controlling, tolgyessy2017foundations}, health monitoring systems \cite{jung2015wearable}, interactive games \cite{park2006vision} and sign language recognition \cite{neverova2014}.

The aim of our work is to develop a robust, vision-based gestures recognition strategy suitable for human-robot/computer interaction tasks in social or industrial settings.
Industrial applications where human safety is critical, often require specialized sensors compatible with safety standards such as ISO/TS 15066.
Yet, scenarios which require human sensing in industry or in social settings are broad. Monocular cameras offer benefits which specialized or multi-modal sensors do not have, such as being lightweight, inexpensive, platform independent and easy to integrate.
This is desirable for robotic assistants in commercial businesses such as restaurants, hotels, or clinics.
We therefore, propose a unified
framework for recognizing static and dynamic gestures from RGB images/video sequences.

A study of gestural communication~\cite{Gleeson2013a} notes that most gestures used in assembly tasks are physically simple while no non-hand body language is involved in part manipulation.
At first, we design a robust static hand gestures detector which is trained on a background substituted gestures dataset namely \textit{OpenSign} \cite{OpenSign_mazhar}, which contains 9 American Sign Language (ASL) gestures.
Sign language is considered among the most structured set of gestures \cite{Starner1998a}.
In this work, we employ sign language gestures only as a proof of concept: our static hand gesture detector can be adapted to other classes as well.
The static hand gestures detector is detailed in \cite{mazhar2019real}.
For more generic and flexible gesture detection, we propose a multi-stream neural architecture for dynamic gestures recognition, which is integrated with our static hand gestures detector in a unified network.

Our unified network is named \textit{StaDNet~\textendash~}Static and Dynamic gestures Network.
It learns to smartly focus on dominant input stream(s) to correctly recognize
large-scale upper-body motions plus subtle hand movements, and therefore distinguish several inter-class ambiguities.
The idea of \textit{visual attention} presented in \cite{rensink2000dynamic} is also embedded in \textit{StaDNet}, which is eventually based the on human selective focus and perception.
Thus, we develop a pose-driven hard spatial-attention mechanism, which focuses on the human upper body and on his/her hands (see Figure~\ref{overall_pipeline}).
It is also noteworthy that in the RGB images, scale information about the subjects (\textit{e.g.}, size of his/her body parts) is lost.
To address this problem, we devise novel \textit{learning-based depth estimators} to regress the distance of the hands and the upper-body from the sensor.
Our depth estimators are trained on the ground truth depth obtained from the video sequences of Kinect V2.
Once the parameters are learned, our algorithm is able to regress the relative depth of the body joints only from the 2D human skeleton.
Therefore, in practice, we no longer require a depth sensor and \textit{StaDNet} is able to detect static and dynamic gestures exclusively from the color images.
This characteristic makes \textit{StaDNet} suitable for inexpensive human-robot interaction in social or industrial settings.

\section{Related Work} \label{literature}
The gestures detection techniques can be mainly divided into two categories: wearable strategies and non-wearable methods.
The wearable strategies include electronic/glove-based systems \cite{Neto2013a, Wong2021a}, and markers-based vision \cite{zhu2011motion} methods.
However, these are often expensive, counter-intuitive and limit the operator's dexterity in his/her routine tasks.
Conversely, non-wearable strategies such as pure-vision based methods, do not require structuring the environment and/or the operator, while they offer ease-of-use to interact with the robots/machines.
Moreover, the consumer-based vision sensors have rich output, are portable and low cost, even when depth is also measured by the sensor such as Microsoft Kinect or Intel Realsense cameras.
Therefore, in this research we opt for a pure vision-based method and
review only the works with vision-based gestures detection.

Traditional activity recognition approaches aggregate local spatio-temporal information via hand-crafted features.
These visual representations include the Harris3D detector~\cite{laptev2005space}, the Cuboid detector~\cite{dollar2005behavior}, dense sampling of video blocks~\cite{Wang2009}, dense trajectories~\cite{Wang2011} and improved trajectories~\cite{Wang2013}.
Visual representations obtained through optical flow, \textit{e.g.},  Histograms of Oriented Gradients (HOG), Histograms of Optical Flow (HOF) and Motion Boundary Histograms (MBH) also achieved excellent results for video classification on a variety of datasets~\cite{Wang2009, wang2016robust}.
In these approaches, global descriptors of the videos are obtained by encoding the hand-crafted features using Bag of Words (BoW) and Fischer vector encodings~\cite{sanchez2013image}.
Subsequently, the descriptors are assigned to one or several nearest elements in a vocabulary~\cite{kantorov2014efficient} while the classification is typically performed through Support Vector Machines (SVMs).
In~\cite{wu2012one}, the authors segmented the human silhouettes from the depth videos using Otsu’s method of global image threshold~\cite{otsu1979threshold}.
They extracted a single Extended-Motion History Image (Extended-MHI) as a global representation for each gesture.
Subsequently, maximum correlations coefficient was utilized to recognize gestures in a One-Shot learning setting.
Other works that utilized One-Shot Learning for gesture recognition include~\cite{fanello2013keep, konevcny2014one, wan2013one}.

Lately, the tremendous success of deep neural networks on image classification tasks~\cite{he2016deep, simonyan2014very} instigated its application in activity recognition domain.
The literature on the approaches that exploit deep neural networks for gestures/activity recognition is already enormous.
Here, we focus on related notables which have inspired our proposed strategy.

\subsection{3D Convolutional Neural Networks}
Among the pioneer works in this category,~\cite{ji20123d} adapted Convolutional Neural Networks (CNNs) to 3D volumes (3D-CNNs), obtained by stacking video frames, to learn spatio-temporal features for action recognition.
In~\cite{baccouche2011sequential}, Baccouche et al. proposed an approach for learning the evolution of temporal information through a combination of 3D-CNNs and LSTM recurrent neural networks~\cite{hochreiter1997long}. 
The short video clips of approximately 9 successive frames were first passed through a 3D-CNN features extractor while the extracted features were subsequently fed to the LSTM network.
However, Karpathy et al. in \cite{Karpathy2014} found that the stacked-frames architecture performed similar to the one with single-image input.

A Few-Shot temporal activity detection strategy is proposed in~\cite{xu2018similarity}, which utilized 3D-CNNs for features extraction from the untrimmed input video as well as from the few-shot examples.
A two-stage proposal network was applied on top of the extracted features while the refined proposals were compared using cosine similarity functions.

To handle resulting high-dimensional video representations, the authors of~\cite{zheng2018deep} proposed the use of random projection-based ensemble learning in deep networks for video classification.
They also proposed rectified linear encoding (RLE) method to deal with redundancy in the initial results of the classifiers.
The output from RLE is then fused by a fully-connected layer that produced the final classification results.

\subsection{Multi-modal Multi-scale strategies}
The authors of~\cite{neverova2014} presented a multi-modal multi-scale detection strategy for \textit{dynamic poses} of varying temporal scales as an extension to their previous work \cite{neverova2013multi}.
They utilized the RGB and depth modalities, as well as the articulated pose information obtained through the depth map.
The authors proposed a complex learning method which included pre-training of individual classifiers on separate channels and iterative fusion of all modalities on shared hidden and output layers.
This approach involved recognizing 20 categories from Italian conversational gestures, performed by different people and recorded with an RGB-D sensor.
This strategy was similar in function to \cite{Karpathy2014} except that it included depth images and pose as additional modalities.
However, it lacked a dedicated equipment to learn evolution of temporal information and may fail when understanding long-term dependencies of the gestures is required.

In \cite{miao2017multimodal}, authors proposed a multi-modal large-scale gesture recognition scheme on the \textit{Chalearn 2016 Looking at People Isolated Gestures recognition} dataset \cite{wan2016chalearn}.
In~\cite{tran2017convnet}, ResC3D network was exploited for feature extraction, and late fusion combined features from multi-modal inputs in terms of canonical correlation analysis.
The authors used linear SVM to classify final gestures.
They proposed a \textit{key frame attention mechanism}, which relied on movement intensity in the form of optical flow, as an indicator for frame selection.

\subsection{Multi-stream Optical Flow-based Methods}
The authors of \cite{simonyan2014two} proposed an optical flow-based method exploiting convolutional network networks for activity recognition along the same lines of~\cite{Karpathy2014}.
They presented the idea of decoupling spatial and temporal networks.
The proposed architecture in \cite{simonyan2014two} is related to the two-stream hypothesis of the human visual cortex \cite{goodale1992separate}.
The spatial stream in there work operated on individual video frames, while the input to the temporal stream was formed by stacking optical flow displacement fields between multiple consecutive frames.

The authors of \cite{wang2015action} presented improved results in action recognition, by employing a trajectory-pooled two-stream CNN inspired by \cite{simonyan2014two}. They exploited the concept of improved trajectories as low level trajectory extractor.
This allowed characterization of the background motion in two consecutive frames through the estimation of the homography matrix taking camera motion into account.
Optical flow-based methods (e.g., the \textit{key frame attention mechanism} proposed in \cite{miao2017multimodal}) may help emphasizing frames with motion, but are unable to differentiate motion caused by irrelevant objects in the background. 

\subsection{CNN-LSTM and Convolutional-LSTM Networks}
The work in~\cite{yue2015beyond} proposed aggregation of frame-level CNN activations through 1) Feature-pooling method and 2) LSTM network for longer sequences.
The authors argued that the predictions on individual frames of video sequences or on shorter clips as performed in \cite{Karpathy2014}, might only contain local information of the video description, while it could also confuse classes if there are fine-grained distinctions.

The authors in~\cite{donahue2015long} proposed a Long-term Recurrent Convolutional Network (LRCN) for multiple situations including sequential input and static output for cases like activity recognition.
The visual features from RGB images were extracted through a deep CNN, which were then fed into stacked LSTM in distinctive configurations corresponding to the task at hand.
The parameters were learned in an ``end-to-end" fashion, such that the visual features relevant to the sequential classification problem were extracted. 

The authors in \cite{xingjian2015convolutional} proposed a method to process sequential images through Convolutional-LSTM (ConvLSTM), which is a variant of LSTM containing a convolution operation inside the LSTM cell.
In \cite{zhu2019redundancy}, the authors studied redundancy and attention in ConvLSTM by deriving its several variants for gesture recognition.
They proposed Gated-ConvLSTM by removing spatial convolutional structures in the gates as they scarcely contributed to the spatio-temporal feature fusion in their study.
The authors evaluated results on the \textit{Chalearn 2016} dataset and found that the Gated-ConvLSTM achieved reduction in parameters size and in computational cost. However, it did not improve detection accuracy to a considerable amount.


\subsection{Multi-Label Video Classification}
The authors of \cite{yeung2018every} presented a multi-label action recognition scheme.
It was based on Multi-LSTM network which tackled with multiple inputs and outputs.
The authors fine-tuned VGG-16 pre-trained on ImageNet \cite{krizhevsky2012imagenet}, on Multi-THUMOS dataset at the individual frame level.
Multi-THUMOS is an extension of THUMOS dataset \cite{idrees2017thumos}.
A fixed length window of 4096-dimensional ``fc7'' features of the fine-tuned VGG-16 was passed as input to the LSTM, through an attention mechanism, that weighted the contribution of individual frames in the window.

\subsection{Attention-based Strategies}
The application of convolutional operations on entire input images tends to be computationally expensive.
In \cite{rensink2000dynamic}, Rensink discussed the idea of \textit{visual representation}, which implied that the humans do not form detailed depiction of all objects in a scene.
Instead, their perception focuses selectively on the objects needed immediately.
This was supported by the concept of \textit{visual attention} applied for deep learning methods as in \cite{mnih2014recurrent}.

Baradel et al.~\cite{baradel2017pose} proposed a spatio-temporal attention mechanism conditioned on human pose. The proposed spatial-attention mechanism was inspired by the work of Mnih et al.~\cite{mnih2014recurrent} on glimpse sensors.
A spatial attention distribution was learned conjointly through the hidden state of the LSTM network and through the learned pose feature representations.
Later, Baradel et al. extend their work in~\cite{baradel2017human} and proposed that the spatial attention distribution can be learned only through an \textit{augmented pose vector}, which was defined by the concatenation of current pose, velocity and accelerations of each joint over time.

The authors in \cite{zheng2020global} proposed a three streams attention network for activity detection.
These were statistic-based, learning-based and global-pooling attention streams.
Shared ResNet was used to extract spatial features from image sequences.
They also proposed a global attention regularization scheme to enable the employed recurrent networks to learn dynamics based on global information.

Lately, the authors of~\cite{narayana2018gesture} presented the state-of-the-art results on the~\textit{Chalearn 2016} dataset. They proposed a novel multi-channel architecture, namely FOANet, built upon a spatial \textit{focus of attention (FOA)} concept.
They cropped the regions of interest occupied by the hands in the RGB and depth images, through the region proposal network and Faster R-CNN method.
The architecture comprised of 12 channels in total with: 1 global (full-sized image) channel and 2 focused (left and right hand crops) channels for each of the 4 modalities (RGB, depth and optical flow fields extracted from the RGB and depth images).
The softmax scores of each modality were fused through a sparse fusion network.

\begin{figure*}[ht]
	\centering
	\includegraphics[width=0.9\linewidth, trim={0.55cm 0.2cm 1.3cm 0.3cm},clip]{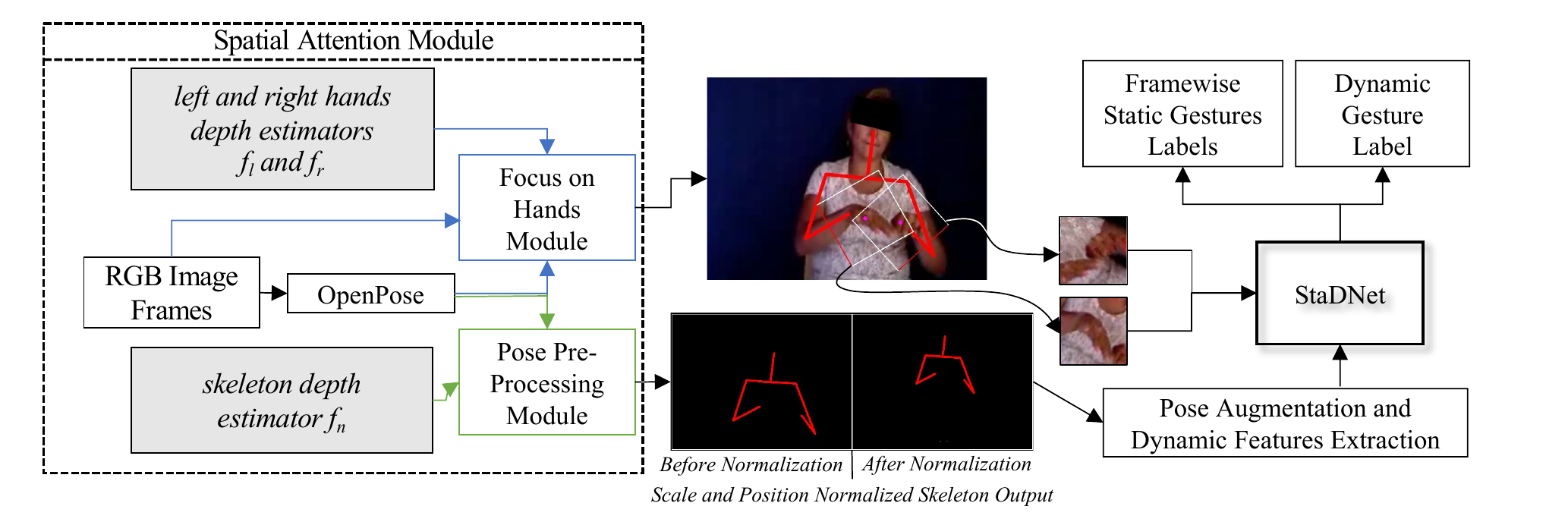}
	\caption{Illustration of our proposed framework.
		In Spatial Attention Module, we mainly have learning-based depth estimators (grey boxes), Focus on Hands (FOH) Module and Pose Pre-Processing (PP) Module.
		2D skeleton is extracted by \textit{OpenPose}.
		FOH exploits hand coordinates obtained from the skeleton and crops hand images with the help of hand depth estimators,
		while PP performs scale and position normalization of the skeleton with the help of skeleton depth estimator.
		The features from the normalized pose are extracted by Pose Augmentation and Dynamic Features Extraction Module and are fed to \textit{StaDNet} together with the cropped hand images.
		\textit{StaDNet} detects frame-wise static gestures as well as dynamic gestures in each video.
	}
	\label{overall_pipeline}
\end{figure*}

\section{Datasets}\label{datasets}

For dynamic gestures classification, we use the \textit{Chalearn 2016 Isolated Gestures dataset} \cite{wan2016chalearn}, referred to simply as \textit{Chalearn 2016} in the rest of the paper.
It is a large-scale dataset which contains Kinect V1 color and depth recordings in $320 \times 240$ resolution of $249$ dynamic gestures recorded with the help of $21$ volunteers.
The gestures vocabulary in the \textit{Chalearn 2016} is mainly from nine groups corresponding to the different application domains: body language gestures, gesticulations, illustrators, emblems, sign language, semaphores, pantomimes, activities and dance postures.
The dataset has $47,930$ videos with each video (color $+$ depth) representing one gesture.
It has to be noted that the Chalearn 2016 does not take into account any specific industry requirements, and that Kinect V1 is obsolete.
However, as we intend to target a broader human-robot interaction domain which includes the fast-growing field of socially assistive as well as household robotics, this requires robots to have the capacity to capture, process and understand human requests in a robust, natural and fluent manner.
Considering the fact that the Chalearn 2016 dataset offers a challenging set of gestures taken from a comprehensive gestures vocabulary with inter-class similarities and intra-class differences, we assumed the Chalearn 2016 suitable for training and benchmarking results of our strategy.

To demonstrate the utility of our approach on a different gesture dataset, we evaluate the performance of our model on the Praxis gesture dataset \cite{negin2018praxis} as well.
This dataset is designed to diagnose \textit{apraxia} in humans, which is a motor disorder caused by brain damage.
This dataset contains RGB (960$\times$540 resolution) and depth (512$\times$424 resolution) images recorded by 60 subjects plus 4 clinicians with Kinect V2.
In total, 29 gestures were performed by the volunteers (15 static and 14 dynamic gestures).
In our work, only dynamic gestures i.e., 14 classes are considered while their pathological aspect is not taken into account i.e., only gestures labeled ``correct'' are selected.
Thus, the total number of considered videos in this dataset is 1247 with mean length of all samples equal to 54 frames.
\textit{StaDNet} is trained exclusively on color images of these datasets for dynamic gestures detection.

\section{Our Strategy}

In this work, we develop a novel unified strategy to model human-centered spatio-temporal dependencies for the recognition of static as well as dynamic gestures.
Our \textit{Spatial Attention Module} localizes and crops hand images of the person, which are subsequently passed as inputs to \textit{StaDNet} unlike previous methods that take entire images as input e.g., \cite{donahue2015long, yue2015beyond}.
Contrary to \cite{yeung2018every}, where a pre-trained state-of-the-art network is fine-tuned on entire image frames of gestures datasets, we fine-tune \textit{Inception V3} on a background-substituted hand gestures dataset, used as our CNN block.
Thus, our CNN has learned to concentrate on image pixels occupied exclusively by hands. This enables it to accurately distinguish subtle hand movements.
We have fine-tuned \textit{Inception V3} with a \textit{softmax} layer, to classify 10 ASL static hand gestures while the features from the last fully connected (FC) layer of the network are extracted as image-embeddings of size $1024$ elements.
These are used as input to the dynamic gestures detector in conjunction with the augmented pose vector which we explain in Sections~\ref{section_scale_norm} and \ref{dynamic_features}.
Moreover, in contrast to the previous strategies for dynamic gestures recognition/video analysis \cite{neverova2014,baradel2017pose,baradel2017human} which employed 3D human skeletons to learn large-scale body motion~\textendash~and corresponding sensor modalities~\textendash~we only utilize 2D upper-body skeleton as an additional modality to our algorithm.
However, scale information about the subjects is lost in monocular images.
To address this, we also propose \textit{learning-based depth estimators}, which determine the approximate depth of the person from the camera and region-of-interest around his/her hands from upper-body 2D skeleton coordinates only.
In a nutshell, \textit{StaDNet} only exploits the RGB hand images and an augmented pose vector obtained from 8 upper-body 2D skeleton coordinates, unlike other existing approaches like \cite{narayana2018gesture}, which include full-frame images in addition to hand images, depth frames and even optical flow frames altogether.

To reiterate, our  method  does  not  require  depth  sensing.
We only utilized the (raw) depth map from Kinect V2 offline, to obtain ground truth depth values of a given 2D skeleton for our \textit{learning-based depth estimators}.
These values can be obtained from any state-of-the-art depth sensor.
Once the depth estimators are trained, our method only requires RGB modality to process images and detect gestures on-line.
We employ~\textit{OpenPose}~\cite{cao2017realtime} which is an efficient discriminative 2D pose extractor, to extract the human skeleton and human hands' keypoints in images. 
\textit{OpenPose} also works exclusively on the RGB images.
Thus, our method can be deployed on a system with any RGB camera, be it a webcam or an industrial color (or RGB-D) camera.
Nevertheless, we only tested \textit{OpenPose} in laboratory or indoor domestic environments and not in a real industry.
Yet, since our framework is not restricted to the use of \textit{OpenPose}, we could integrate another pose extractor system, better suited for the target application scenario.

\begin{figure*}[t]
	\centering
	\includegraphics[width=0.99\linewidth]{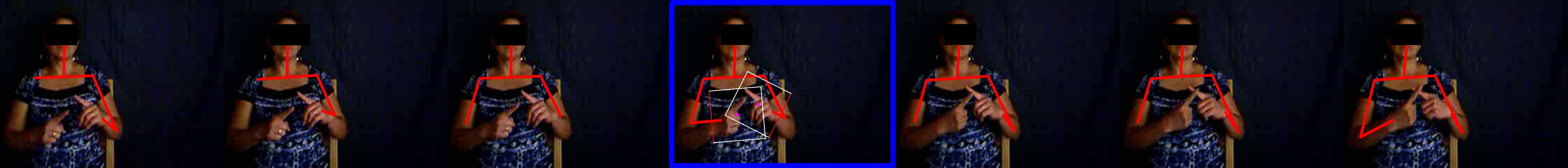}
	\caption{The \textit{Skeleton Filter} described in Section~\ref{filtering_skeleton}. Images are arranged from left to right in chronological order. The central image shows the skeleton output by the filter. The six other images show the raw skeletons output by \textit{OpenPose}. Observe that -- thanks to equation~(\ref{eqn_replace_one_non_zero}) -- our filter has added the right wrist coordinates (shown only in the central image). These are obtained from the $K$-th frame, while they were missing in all raw skeletons from frame $1$ to $K-1$.
	}
	\vspace{-1em}
	\label{image_skeleton_filtering}
\end{figure*}

\section{Spatial Attention Module} \label{spatial_attention}
Our spatial attention module is divided into two parts: \textit{Pose Pre-processing Module} and \textit{Focus on Hands Module} (see Figure~\ref{overall_pipeline}). We detail these modules in the following.

\subsection{Pose Pre-processing Module} \label{pose_extraction}

We first resize the dataset videos to $1080 \times C$ pixels, where $C$ is the value of resized image columns obtained with respect to new row value i.e., 1080, while maintaining the aspect ratio of the original image (1440 in our work).
The necessity to resize the input videos will be explained in Section \ref{depth_estimation}. 
After having resized the videos, we feed them to \textit{OpenPose}, one at a time, and the output skeleton joint and hand keypoint coordinates are saved for offline pre-processing. 
The pose pre-processing is composed of three parts, detailed hereby: \textit{skeleton filter}, \textit{skeleton position and scale normalization} and \textit{skeleton depth estimation}.

\subsubsection{Skeleton Filter} \label{filtering_skeleton}
For each image, \textit{OpenPose} extracts $N$ skeleton joint coordinates depending on the selected body model while it does not employ pose tracking between images. The occasional jitter in the skeleton output and missing joint coordinates between successive frames may hinder gesture learning. Thus, we develop a \textit{two-step} pose filter that rectifies occasional disappearance of the joint(s) coordinates and smooths the \textit{OpenPose} output. The filter operates on a window of $K$ consecutive images ($K$ is an adjustable odd number, $7$ in this work), while the filtered skeleton is obtained in the center frame.
We note $\mathbf{p}^i_k = (x^{i}, y^{i})$, the image coordinates of the $i_{th}$ joint in the skeleton output by \textit{OpenPose} at the $k$-th image within the window. If \textit{OpenPose} does not detect joint $i$ on image $k$: $\mathbf{p}^i_k = \emptyset$. 

In a \textit{first step}, we replace coordinates of the missing joints.
Only $\bar{r}$ (we use $\bar{r} = 7$) consecutive replacements are allowed for each joint $i$, and we monitor this via a coordinate replacement counter, noted $r^i$. The procedure is driven by the following two equations:
\begin{equation} 
\begin{array}{ll}
\mathbf{p}^i_K = \mathbf{p}^i_{K-1} & \text{ if } \mathbf{p}^i_K = \emptyset\\
& \wedge  \; \mathbf{p}^i_k \neq \emptyset\;  \forall k=1,\dots,K-1 \\
& \wedge \; r^i \leq \bar{r}
\label{eqn_replace_one_non_zero}
\end{array}
\end{equation} 
\begin{equation} 
\begin{array}{ll}
\mathbf{p}^i_{k=1,\dots,K-1} = 
\begin{cases}
\emptyset & \text{if } \mathbf{p}^i_K = \emptyset \wedge
r^i > \bar{r}\\
\mathbf{p}^i_K& \text{if } \mathbf{p}^i_{k=1,\dots,K-1} = \emptyset \wedge \mathbf{p}^i_K \neq \emptyset
\end{cases}
\label{eqn_replace_with_zero_or_all_non_zero}
\end{array}
\end{equation}
Equation~(\ref{eqn_replace_one_non_zero}) states that the i-th joint at the latest (current) image $K$ is replaced by the same joint at the previous image $K-1$ under three conditions: if it is not detected, if it has been detected in all previous images, and if in the past it has not been replaced up to $\bar{r}$ consecutive times already.
If any of the conditions is false, we do not replace the coordinates and we reset the replacement counter for the considered joint: $r^i=0$. Similarly,~(\ref{eqn_replace_with_zero_or_all_non_zero}) states that the i-th joint coordinates over the window should not be taken into account i.e., joint will be considered missing, if it is not detected in the current image $K$ and if it has already been replaced more than $\bar{r}$ consecutive times (we allow only $\bar{r}$ consecutive replacements driven by~(\ref{eqn_replace_one_non_zero})). This also resets the replacement counter value for the considered joint. Moreover, the i-th joint in all of the window's $K-1$ images is set to its position in the current image $K$, if it has never been detected in the window up to the current image.

In \textit{second step}, we apply Gaussian smoothing to each $\mathbf{p}^i$, over the window of $K$ images. Applying this filter removes jitter from the skeleton pose and smooths out the joint movements in the image at the center of the filter window.
Figure~\ref{image_skeleton_filtering} shows the output of our skeleton filter for one window of images.


\subsubsection{Skeleton Position and Scale Normalization}\label{section_scale_norm}
\begin{figure}[H]
	\centering
	\includegraphics[width=0.99\linewidth]{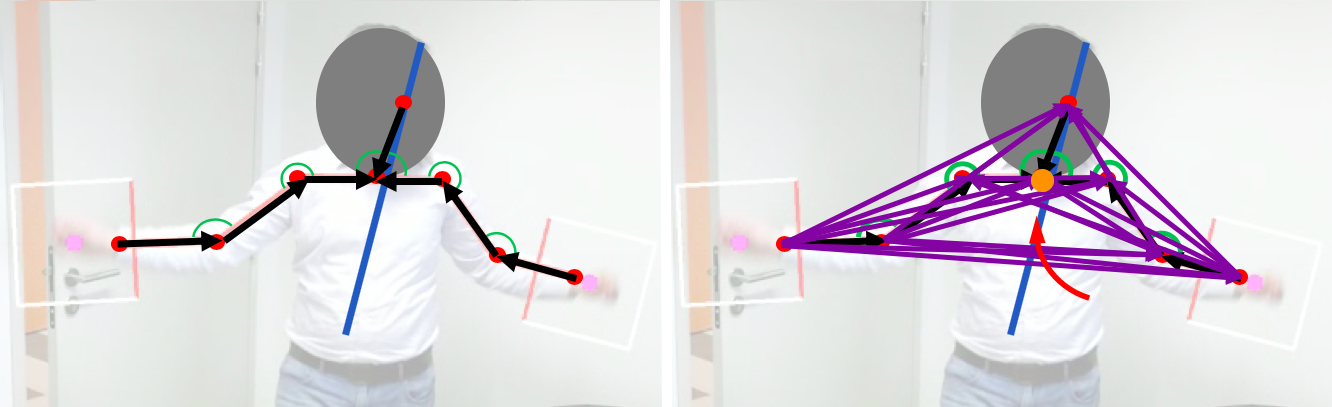}
	\caption{Features augmentation of the upper body. 
	In the left image, we show 8 upper-body joint coordinates (red), vectors connecting these joints (black) and angles between these vectors (green). From all upper-body joints, we compute a \textit{line of best fit} (blue). In the right image, we show all the vectors (purple) between unique pairs of upper-body joints.
	We also compute the angles (not shown) between the vectors and the line of best fit.
	From 8 upper-body joints, we obtain 97 components of the \textit{augmented pose vector}.
	}
	\label{scale_normalization}
\end{figure}

Figure~\ref{overall_pipeline} includes a simple illustration of our goal for skeleton position and scale normalization. We focus on the 8 upper-body joints shown in Figure~\ref{scale_normalization}: $\mathbf{p}^{0, \dots,7}$, with $\mathbf{p}^0$ corresponding to the \textit{Neck} joint, which we consider as root node.
\textit{Position normalization} consists in eliminating the influence of the user's position in the image, by subtracting the \textit{Neck} joint coordinates from those of the other joints. \textit{Scale normalization} consists in eliminating the influence of the user's depth. We do this by dividing the position-shifted joint coordinates by the neck depth $d_n$, on each image, so the all joints are replaced according to: 
\begin{equation}
\mathbf{p}^i \leftarrow \frac{\mathbf{p}^i-\mathbf{p}^0}{d_n}.     
\label{eq:norm}
\end{equation}
Since our framework must work without requiring a depth sensor, we have developed a skeleton depth estimator to derive the neck depth, $\widetilde{d}_n$ and use it instead of $d_n$ in~(\ref{eq:norm}).
This estimator is a neural network, which maps a 97-dimensional pose vector, derived from the 8 upper body joint positions, to the depth of the \textit{Neck} joint. We will explain it hereby.

\subsubsection{Skeleton Depth Estimation}\label{depth_estimation}
Inspired by~\cite{neverova2014}, which demonstrated that augmenting pose coordinates may improve performance of gesture classifiers, we develop a 97 dimensional \textit{augmented pose vector} $\mathbf{x_n}$ (subscript \textbf{n} means \textit{Neck} here) from 8 upper-body joint coordinates.
From the joints coordinates, we obtain --  via least squares -- a \textit{line of best fit}.
In addition to 7 vectors from anatomically connected joints, 21 vectors between unique pairs of all upper-body coordinates are also obtained.
The lengths of individual augmented vectors are also included in $\mathbf{x_n}$. We also include the 6 angles formed by all triplets of anatomically connected joints, and the 28 angles, between the 28 (anatomically connected plus augmented) vectors and the \textit{line of best fit}.
The resultant 97-dimensional augmented pose vector concatenates: 42 elements from abscissas and ordinates of the augmented vectors, their 21 estimated lengths and 34 relevant angles.

To obtain the ground-truth depth of \textit{Neck} joint, denoted $d_n$, we utilize \textit{OpenSign} dataset.
\textit{OpenSign} is recorded with Kinect V2 which outputs the RGB and the registered depth images with resolution 1080$\times$1920.
We apply our augmented pose extractor to all images in the dataset and -- for each image -- we associate $\mathbf{x_n}$ to the corresponding \textit{Neck} depth.
A 9 layers neural network $f_n$ is then designed, to optimize parameters $\theta_n$, given augmented pose vector $\mathbf{x_n}$ and ground-truth $d_n$ to regress the approximate distance value $\widetilde{d}_n$ with a mean squared error of $8.34 \times 10^{-4}$. 
Formally:
\begin{flalign}\label{scale_norm_deep_model}
\widetilde{d}_n = f_n(\mathbf{x_n}, d_n; \theta_n).
\end{flalign}
It is to be noted that the estimated  depth $\widetilde{d}_n$ is  a relative  value  and not  in metric  units,  and  that the resolution  of ground truth images in \textit{OpenSign} is 1080$\times$1920.
For scale normalization (as explained in Section~\ref{section_scale_norm}), we utilize the estimated  depth $\widetilde{d}_n$.
Thus, the input images from the Chalearn 2016 dataset are resized such that the row count of the images is maintained to 1080.
This is required as we need to re-scale the predicted depth to the original representation of the depth map in \textit{OpenSign} (or to that of Kinect V2).
Yet, the \textit{StaDNet} input image size can be adapted to the user’s needs if the depth estimators are not employed.

\subsection{Focus on Hands Module} \label{focus_on_hands}
This module focuses on hands in two steps: first, by localizing them in the scene, and then by determining the size of their bounding boxes, in order to crop hand images.

\subsubsection{Hand Localization}

One way to localize hands in an image is to exploit Kinect SDK or middleware like OpenNI (or its derivatives).
These libraries however do not provide accurate hand-sensing and are deprecated as well.
Another way of localizing hands in an image is via detectors, possibly trained on hand images as in \cite{panteleris2018using}.
Yet, such strategies struggle to distinguish left and right hands, since they operate locally, thus lacking contextual information.
To keep the framework generic, we decided not to employ specific  hand sensing functionalities from  Kinect – be it V1 or V2 – or other more modern sensing devices.
Instead, we localize the hand via the hand key-points obtained from \textit{OpenPose}.
This works well for any RGB camera and therefore does not require a specific platform (e.g., Kinect) for hand sensing.

\textit{OpenPose} outputs 42 (21 per hand) hand key-points on each image.
We observed that these key-points are more susceptible to jitter and misdetections than the skeleton key-points, particularly on the low resolution videos of the \textit{Chalearn 2016} dataset.
Therefore, we apply the same filter of equations (\ref{eqn_replace_one_non_zero}) and (\ref{eqn_replace_with_zero_or_all_non_zero}) to the raw hand key-points output by \textit{OpenPose}. Then, we estimate the mean of all $N_j$ detected hand key-point coordinates $\mathbf{p}^j$, to obtain:
\begin{flalign} \label{eqn_hand_center_point}
\mathbf{p}^{c} = \frac{1}{N_j}\sum_{j=1}^{N_j} \mathbf{p}^{j},
\end{flalign}
the hand center in the image.

\subsubsection{Hand Bounding-box Estimation} \label{hands}
Once the hands are located in the image, the surrounding image patches must be cropped for gesture recognition. 
Since at run-time our gestures recognition system relies only on the RGB images (without depth), we develop two additional neural networks, $f_l$ and $f_r$, to estimate each hand's bounding box size. These networks are analogous to the one described in Section~\ref{section_scale_norm}.
Following the scale-normalization approach, for each hand we build a 54 dimensional augmented pose vector from 6 key-points.
These augmented pose vectors ($\mathbf{x_l}$ and $\mathbf{x_r}$) are mapped to the ground-truth hands depth values ($d_l$ and $d_r$) obtained from \textit{OpenSign} dataset, through two independent neural networks:
\begin{flalign}\label{left_bounding}
&\widetilde{d}_l = f_l(\mathbf{x_l}, d_l; \theta_l) \\ \label{right_bounding}
&\widetilde{d}_r = f_r(\mathbf{x_r}, d_r; \theta_r).
\end{flalign}
In~(\ref{left_bounding}) and~(\ref{right_bounding}), $f_l$ and $f_r$ are 9-layer neural networks that optimize parameters $\theta_l$ and  $\theta_r$ given augmented poses $\mathbf{x_l}$ and $\mathbf{x_r}$ and ground-truth depths $d_l$ and $d_r$, to estimate depths $\widetilde{d}_l$ and $\widetilde{d}_r$.
Mean squared error for $f_l$ and $f_r$ are $4.50 \times 10^{-4}$ and  $6.83 \times 10^{-4}$ respectively.
The size of the each bounding box is inversely proportional to the corresponding depth ($\widetilde{d}_l$ or $\widetilde{d}_r$) obtained by applying~(\ref{left_bounding}) to the pure RGB images.
The orientation of each bounding box is estimated from the inclination between corresponding forearm and horizon.
The final outputs are the cropped images of the hands, $\mathbf{i_l}$ and $\mathbf{i_r}$.
Now since our depth estimators $f_n$, $f_l$ and $f_r$ have been trained, we do not require explicit depth sensing either to normalize the skeleton or to estimate the hand bounding boxes. 

\section{Video Data Processing} \label{data_processing}
Our proposed spatial attention module conceptually allows end-to-end training of the gestures. However, we train our network in multiple stages to speed-up the training process (the details of which are given in Section~\ref{training}).
Yet, this requires the videos to be processed step-by-step beforehand.
This is done in four steps i.e, (1) 2D pose-estimation, (2) features extraction, (3) label-wise sorting and zero-padding and (4) train-ready data formulation.
While prior 2D-pose estimation may be considered a compulsory step -- even if the network is trained in an end-to-end fashion -- the other steps can be integrated into the training algorithm.

\subsection{Dynamic Features: Joints Velocities and Accelerations}\label{dynamic_features}

As described in Section~\ref{spatial_attention}, our features of interest for gestures recognition are skeleton and hand images.
The concept of augmented pose for scale-normalization has been detailed in Section~\ref{section_scale_norm}.
For dynamic gestures recognition, velocity and acceleration vectors from 8 upper-body joints, containing information about the dynamics of motion, are also appended to the pose vector $\mathbf{x_n}$ to form a new 129 components augmented pose $\mathbf{x_{dyn}}$.
Inspired by \cite{neverova2014}, joint velocities and accelerations are computed as first and second derivatives of the scale-normalized joint coordinates. At each image $k$:
\begin{flalign}\label{eqn_velocity}
&\dot{\mathbf{p}}^i_k = \mathbf{p}^i_{k+1} - \mathbf{p}^i_{k-1} \\ \label{eqn_acceleration}
&\ddot{\mathbf{p}}^i_k = \mathbf{p}^i_{k+2} + \mathbf{p}^i_{k-2} - 2\mathbf{p}^i_{k}.
\end{flalign}

The velocities and accelerations obtained from~(\ref{eqn_velocity}) and~(\ref{eqn_acceleration}) are scaled by the video frame-rate to make values time-consistent, before appending them in the augmented pose vector $\mathbf{x_{dyn}}$. 
For every frame output by the skeleton filter of Section~\ref{filtering_skeleton}, scale-normalized augmented pose vectors $\mathbf{x_{dyn}}$ (as explained in \ref{section_scale_norm}) plus left $\mathbf{i_l}$ and right $\mathbf{i_r}$ hands cropped images (extracted as explained in Section~\ref{focus_on_hands}) are appended in three individual arrays.

\subsection{Train-Ready Data Formulation}
The videos in the \textit{Chalearn 2016} are randomly distributed. Once the features of interest ($\mathbf{i_l}$, $\mathbf{i_r}$ and $\mathbf{x_{dyn}}$) are extracted and saved in \textit{.h5} files, we sort them with respect to their labels.
It is natural to expect the dataset videos (previously sequences of images, now arrays of features) to be of different lengths.
The average video length in this dataset is 32 frames, while we fix the length of each sequence to 40 images in our work.
If the length of a sequence is less than 40, we pad zeros symmetrically at the start and end of the sequence.
Alternatively, if the length is greater than 40, we perform symmetric trimming of the sequence.
Once the lengths of sequences are rectified (padded or trimmed), we append all corresponding sequences of a gesture label into a single array.
At the end of this procedure, we are left with the 249 
gestures in the \textit{Chalearn 2016} dataset, along with an array of the ground-truth labels.
Each feature of the combined augmented pose vectors is normalized to zero mean and unit variance, while for hand images we perform pixel-wise division by the maximum intensity value (e.g., 255).
The label-wise sorting presented in this section is only necessary if one wants to train a network on selected gestures (as we will explain in Section~\ref{training}).
Otherwise, creating only a ground-truth label array should suffice.

\section{Dynamic Gesture Recognition}\label{model}

To classify dynamic gestures, \textit{StaDNet} learns to model the spatio-temporal dependencies of the input video sequences.
As already explained in Sections~\ref{focus_on_hands}~and~\ref{dynamic_features}, we obtain cropped hand images $\mathbf{i_l}$ and $\mathbf{i_r}$ as well as the augmented pose vector $\mathbf{x_{dyn}}$ for each frame in a video sequence.
These features are aggregated in time through Long-Short Term Memory networks to detect dynamic gestures performed in the videos.
However, we do not pass raw hand images, but extract image embeddings of size $1024$ elements per hand. These image embeddings are extracted from the last fully connected layer of our static hand gesture detector and can be considered as rich latent space representations of hand gestures. This is done according to:
\begin{flalign}\label{embeddings}
\mathbf{{e}_l}, \mathbf{p_{l}} = g_{sta}(\mathbf{i_l}, \theta_{st}) \nonumber \\
\mathbf{{e}_r}, \mathbf{p_{r}} = g_{sta}(\mathbf{i_r}, \theta_{st}),
\end{flalign}
with: 
\begin{itemize}
    \item 
$g_{sta}$ the static hand gesture detector, which returns the frame-wise hand gesture class probabilities $\mathbf{p_{l,r}}$ and the embeddings vectors $\mathbf{e_{l,r}}$ from its last fully connected layers;
    \item
$\theta_{st}$ the learned parameters of $g_{sta}$.
\end{itemize}

\begin{figure}[h]
	\centering
	\includegraphics[width=0.8\linewidth, trim={0cm 0.3cm 0cm 0.5cm},clip]{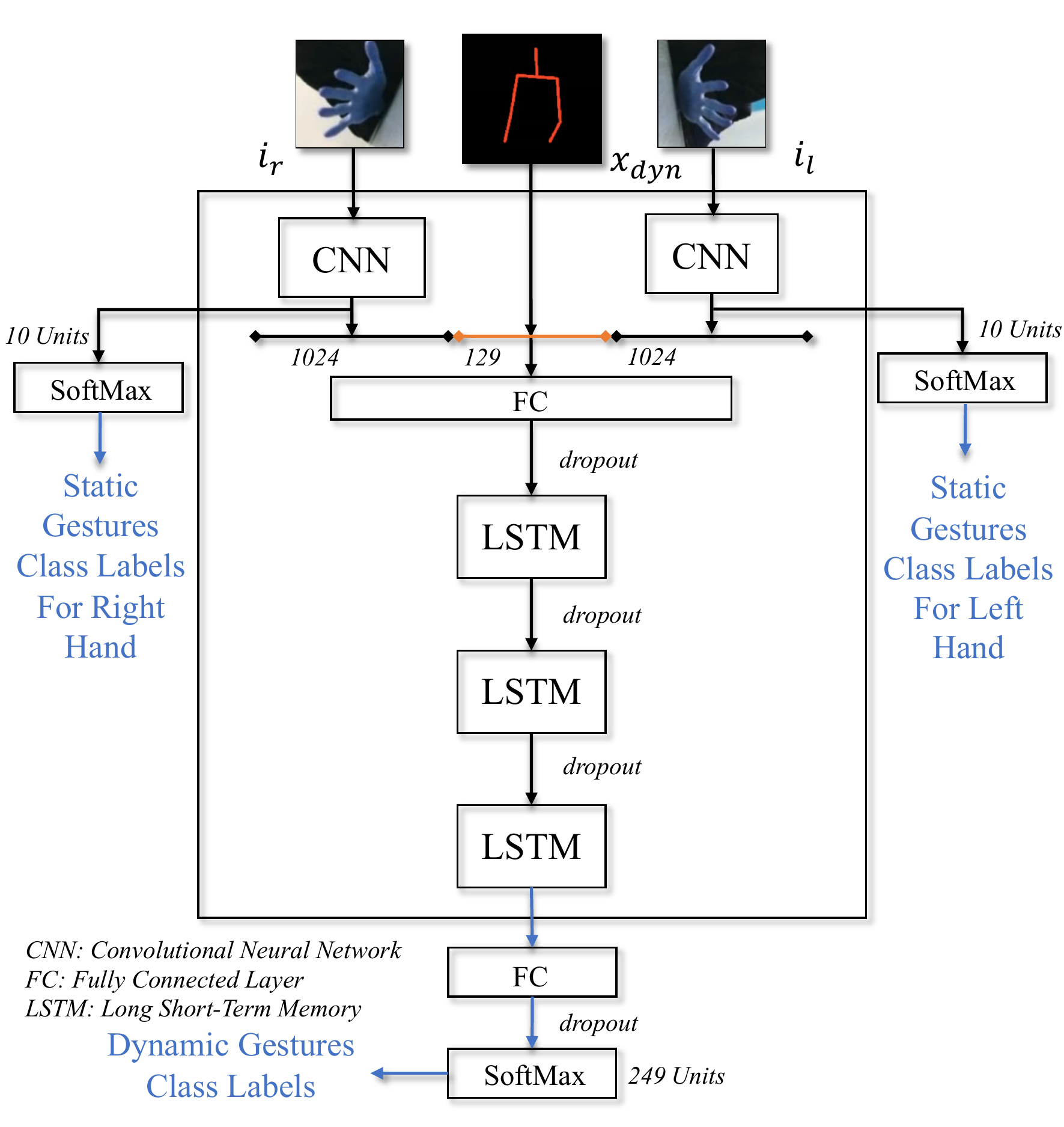}
	\caption{Illustration of \textit{StaDNet} for static and dynamic gestures recognition.
		We perform intermediate fusion to combine hand image embeddings and augmented pose vector.}
	\label{fig_cnn_lstm}
\end{figure}

For each frame of a video sequence of length $N$, the obtained hand image embeddings $\mathbf{e_l}$, $\mathbf{e_r}$ and augmented pose vector $\mathbf{x_{dyn}}$ are subsequently fused in vector ${\psi}$, and then passed to stacked LSTMs followed by $g_{dyn}$ network. This network outputs dynamic gestures probability $p_{dyn}$ for each video:
\begin{gather}
{\psi} = [\mathbf{e_l}; \mathbf{x_{dyn}}; \mathbf{e_r}] \nonumber \\
p_{dyn} = g_{dyn}(LSTMs(\sum_{i=1}^{N} \psi_i, \theta_{LSTMs}), \theta_{dyn}).
\end{gather}
The $g_{dyn}$ network consists of a fully connected layer and a \textit{softmax} layer which takes the output of LSTMs as input; $\theta_{LSTMs}$ and $\theta_{dyn}$ are model parameters to be learned for the detection of dynamic gestures, while $p_{dyn}$ is the detected class probability obtained as output from the \textit{softmax} layer. 
The illustration of our network is presented in Figure~\ref{fig_cnn_lstm}.
We employ dropout regularization method between successive layers to prevent over-fitting and improve generalization, and batch-normalization to accelerate training.

\section{Training} \label{training}
The proposed network is trained on a computer with Intel\textsuperscript{\textcopyright} Core i7-6800K (3.4 GHz) CPU, dual Nvidia GeForce GTX 1080 GPUs, 64 GB system memory and Ubuntu 16.04 Operating system.
The neural network is designed, trained and evaluated in Python - Keras with tensorflow back-end, while skeleton extraction with \textit{OpenPose} is performed in C++.
%

%
The \textit{Chalearn 2016} dataset has 35,875 videos in the provided training set, with only the top 47 gestures (arranged in descending order of the number of samples) representing 34$\%$ of all videos.
The numbers of videos in the provided validation and test sets are 5784 and 6271 respectively. 
The distribution of train, validation and test data in our work is slightly different from the approach proposed in the challenge. 
We combine and shuffle the provided train, validation and test sets together, leading to 47,930 total videos.
For weight initialization, 12210 training videos of 47 gestures are utilized to perform pre-training with a validation split of $0.2$.
We subsequently proceed to train our network for all 249 gestures on 35,930 videos, initializing the parameters with the pre-trained model weights.
In this work, we utilize the \emph{Holdout} cross-validation method, which aligns with the original exercise of the \emph{Chalearn 2016} challenge.
Thus, we optimize the hyper-parameters on the validation data of 6,000 videos, while the results are presented on the test data of the remaining 6000 videos.

As already explained in Section~\ref{datasets}, we utilize only 1247 videos for 14 correctly performed dynamic gestures from the \textit{Praxis Cognitive Assessment Dataset}.
Given the small size of this dataset, we adapt the network hyper-parameters to avoid over-fitting.
\section{Results} \label{results}
For the \textit{Chalearn 2016} dataset, the proposed network is initially trained on 47 gestures with a low learning rate of $1\times10^{-5}$.
After approximately 66,000 epochs, a top-1 validation accuracy of 95.45$\%$ is obtained.
The parameters learned for 47 gestures are employed to initialize weights for complete data training for 249 gestures as previously described.
The network is trained in four phases.
In the first phase, we perform weights initialization, inspired by the \textit{transfer learning} concept of deep networks, by replacing the classification layer (with \textit{softmax} activation function) by the same with output number of neurons corresponding to the number of class labels in the dataset.
In our case, we replace the \textit{softmax} layer in the trained network for 47 gestures plus the FC layer immediately preceding it.
The proposed model is trained for 249 gestures classes with a learning rate of $1 \times 10^{-3}$ and a decay value of $1 \times 10^{-3}$ with \textit{Adam} optimizer.
The early iterations are performed with all layers of the network locked except the newly added FC and \textit{softmax} layers.
As the number of epochs increases, we successively unlock the network layers from the bottom (deep layers).

\begin{figure}[t!]
	\centering
	\includegraphics[width=0.9\linewidth, trim={2cm 1.5cm 2.9cm 9.4cm},clip]{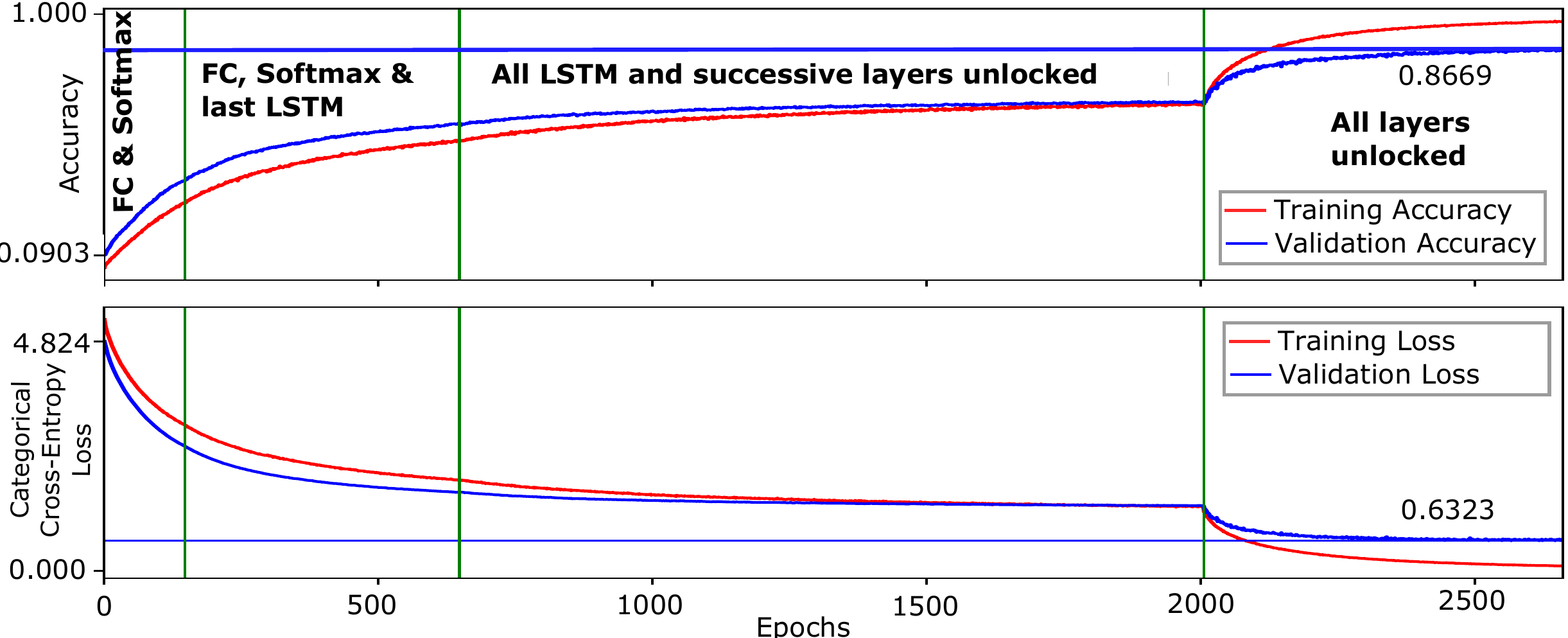}
	\caption{Training curves of the proposed CNN-LSTM network for all 249 gestures of the Chalearn 2016. The network is trained in four phases, distinguished by the vertical lines.}
	\label{train_plot_249}
\end{figure}

\begin{figure}[b!]
	\centering
	\includegraphics[width=0.79\linewidth, trim={0.3cm 4.6cm 0.3cm 4.6cm},clip]{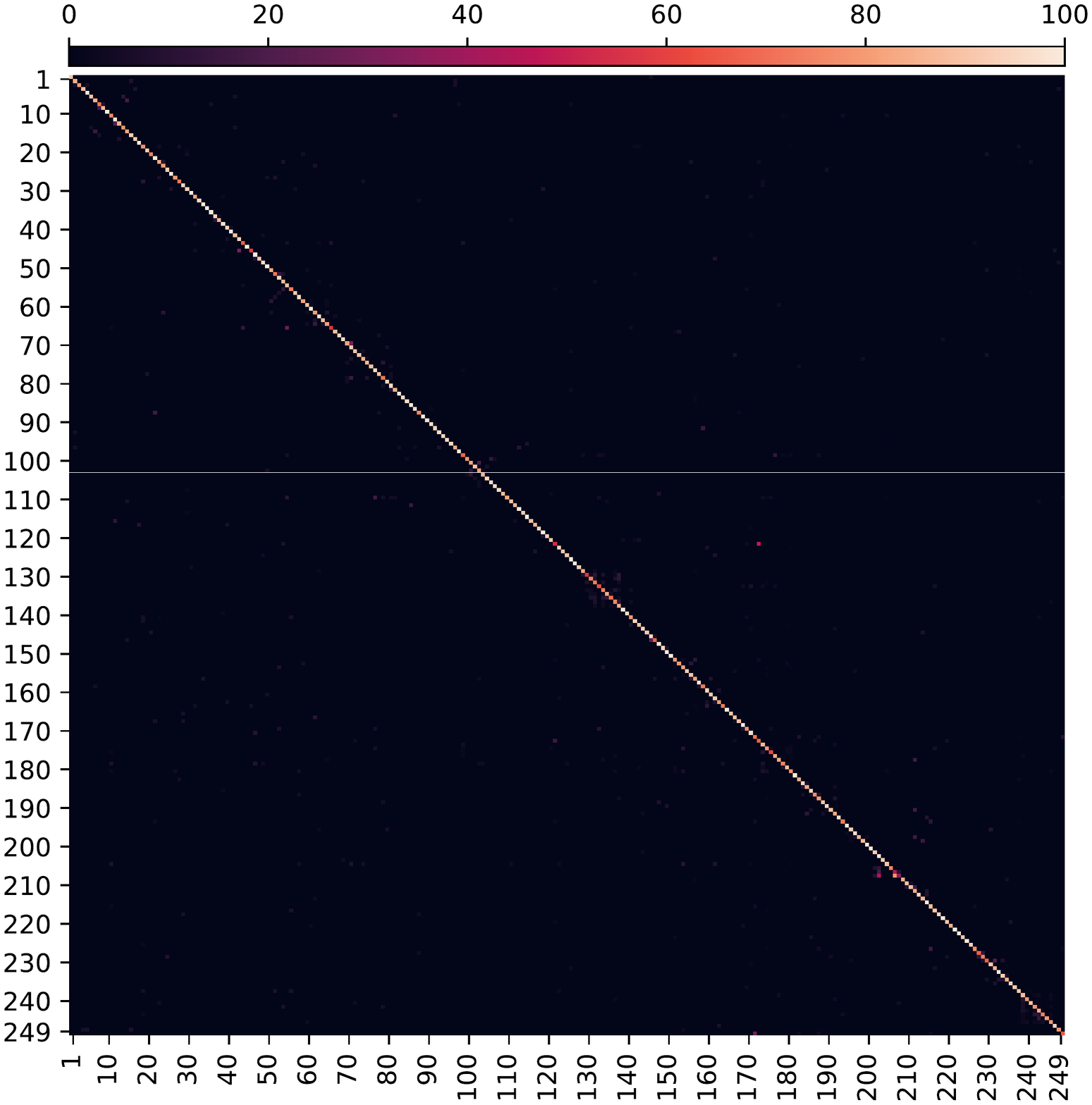}
	\caption{Illustration of the confusion matrix/heat-map of \textit{StaDNet} evaluated on test set of the Chalearn 2016 isolated gestures recognition dataset. It is evident that most samples in the test set are recognized with high accuracy for all 249 gestures (diagonal entries, 86.75$\%$ overall).}
	\label{confusion_matrix_isogd}
\end{figure}

In the second phase, network layers until the last LSTM block are unlocked.
All LSTM blocks and then the complete model are unlocked, respectively in the third and fourth phase.
By approximately 2700 epochs, our network achieves 86.69$\%$ top-1 validation accuracy for all 249 gestures and 86.75$\%$ top-1 test accuracy, surpassing the state-of-art methods on this dataset.
The prediction time for each video sample is 57.17 ms excluding pre-processing of the video frames.
Thus, we are confident that the online dynamic gesture recognition can be achieved in interaction time.
The training curve of the complete model is shown in Figure~\ref{train_plot_249} while the confusion matrix/heat-map with evaluations on test set is shown in Figure~\ref{confusion_matrix_isogd}.
Our results on the \textit{Chalearn 2016} dataset are compared with the reported state-of-the-art in Table~\ref{results_table}.

\begin{table}[H]
	\centering
	\begin{tabular}{|c|c|c|}
		\hline
		\textbf{Method}               & \textbf{Valid $\%$} & \textbf{Test $\%$}  \\ \hline
		\textit{StaDNet} (ours) & \textbf{86.69} & \textbf{86.75} \\ \hline
		FOANet \cite{narayana2018gesture}   & 80.96          & 82.07          \\ \hline
		Miao \textit{et al.} \cite{miao2017multimodal} (ASU)             & 64.40          & 67.71          \\ \hline
		SYSU\_IEEE                    & 59.70          & 67.02          \\ \hline
		Lostoy                        & 62.02          & 65.97          \\ \hline
		Wang \textit{et al.} \cite{wang2017large} (AMRL)            & 60.81          & 65.59          \\ \hline
	\end{tabular}
	\vspace{0.5em}
	\caption{Comparison of the reported results with ours on the Chalearn 2016. The challenge results are published in \cite{wan2017results}. 
	}
	\label{results_table}
\end{table}

Inspecting the training curves, we observe that the network is progressing towards slight over-fitting in the fourth phase when all network layers are unlocked.
Specifically, the first \textit{time-distributed FC} layer is considered the culprit for this phenomenon.
Although we already have a dropout layer immediately after this layer, with dropout rate equaling $0.85$, we skip to further dive deeper to rectify this. However, it is assumed that substitution of this layer with the strategy of \textit{pose-driven temporal attention}~\cite{baradel2017human} or with the \textit{adaptive hidden layer}~\cite{hu2018learning}, may help reduce this undesirable phenomenon and ultimately further improve results.
Moreover, recent studies argue that data augmentation i.e., the technique
of perturbing data without altering class labels, are able to greatly improve model
robustness and generalization performance \cite{Hendrycks2020a}.
As we do not use any data augmentation on the videos in model training for dynamic gestures, doing the contrary might help to reduce over-fitting here.

\begin{figure}[h]
	\centering
	\includegraphics[width=0.8\linewidth, trim={2cm 2cm 2.9cm 2.9cm},clip]{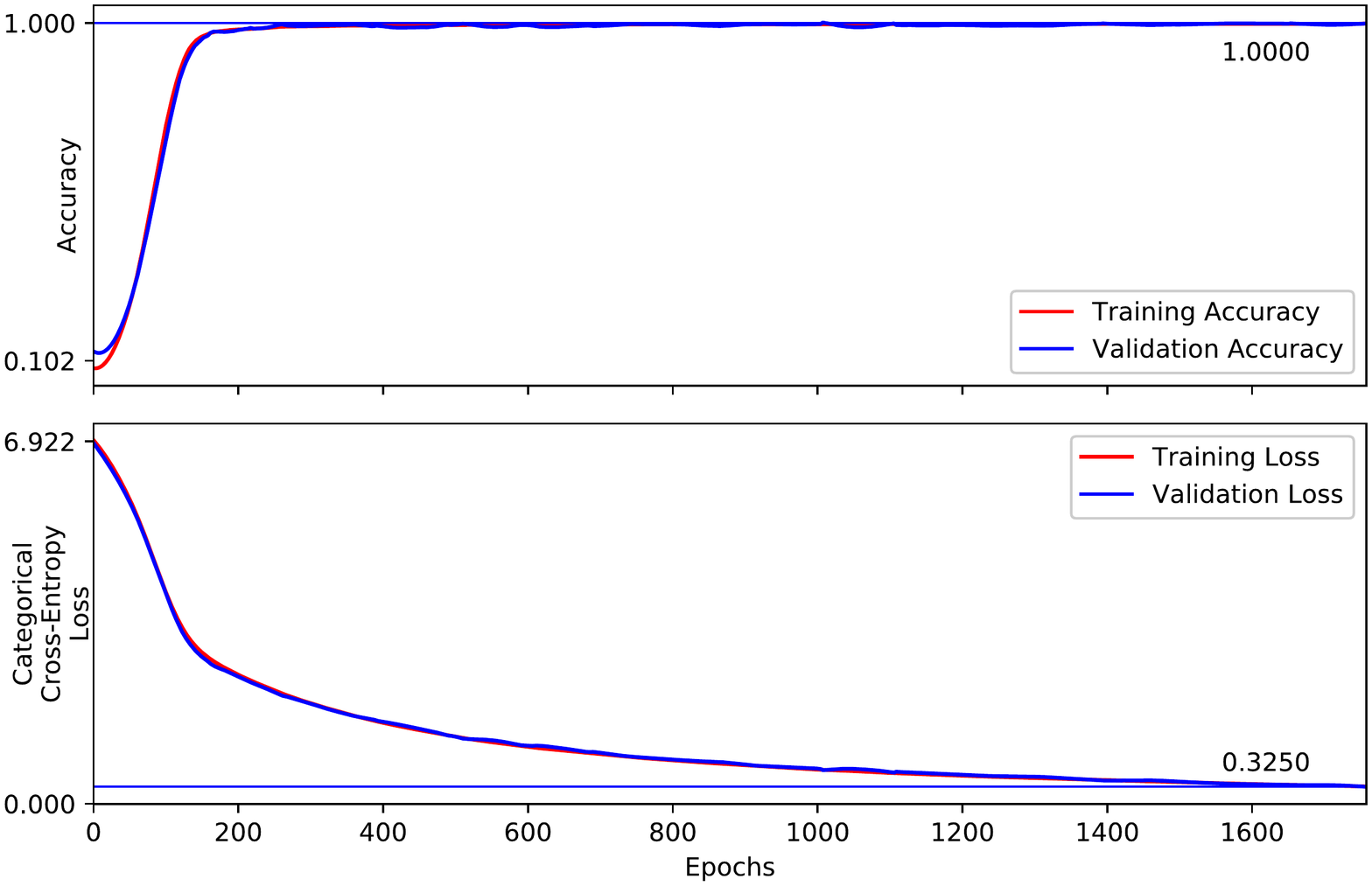}
	\caption{Training curves of \textit{StaDNet} on the Praxis gesture dataset.}
	\label{training_curve_praxis}
\end{figure}

\begin{figure}[t!]
	\centering
	\includegraphics[width=0.85\linewidth, trim={1.5cm 1.5cm 1.6cm 3.35cm},clip]{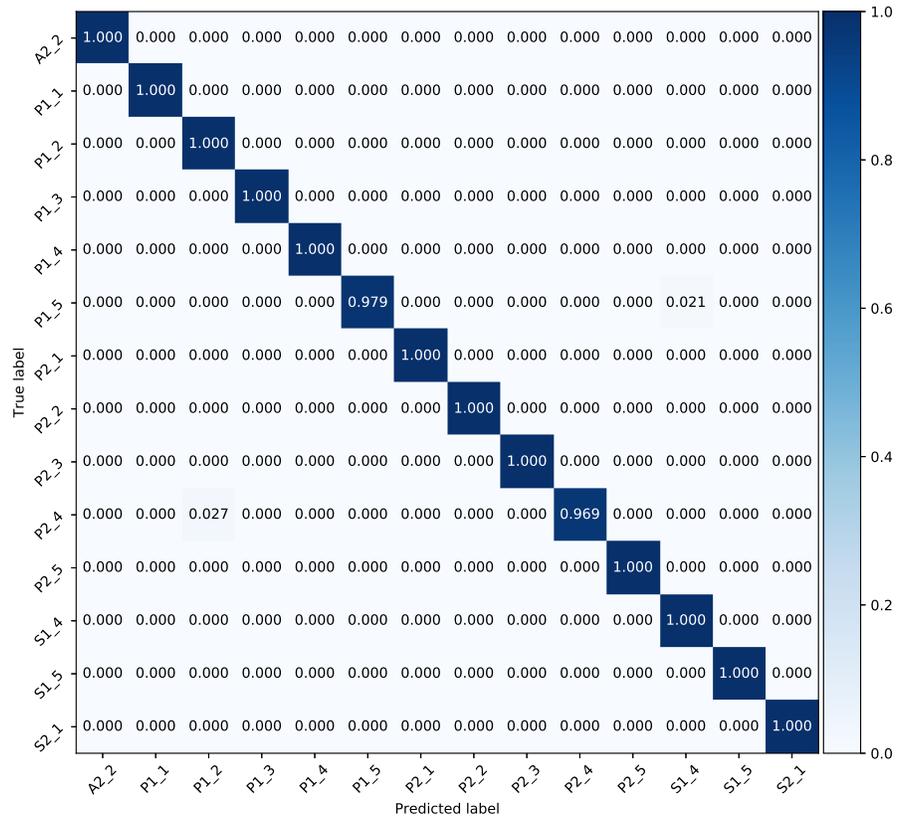}
	\caption{
	Normalized confusion matrix of the proposed model evaluated on test set of the Praxis dataset.
	}
	\label{confusion_matrix_praxis}
\end{figure}

\begin{figure}[t!]
	\centering
	\includegraphics[width=0.92\linewidth, trim={0cm 0.2cm 0cm 0.2cm},clip]{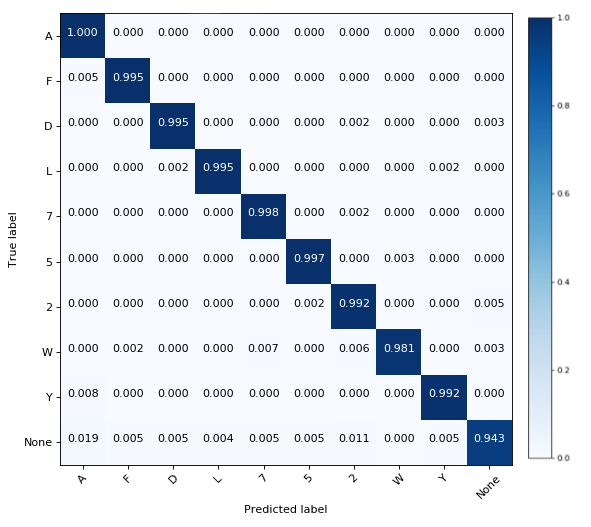}
	\caption{Normalized confusion matrix for our static hand gesture detector quantified on test-set of \textit{OpenSign}. This figure is taken from \cite{mazhar2019real} with the authors' permission.
	}
	\label{confusion_hand}
\end{figure}

For the \textit{Praxis} dataset, the optimizer and values of learning rate and decay, are the same as for the \textit{Chalearn 2016} dataset.
The hyper-parameters including number of neurons in FC layers plus hidden and cell states of LSTM blocks are (reduced) adapted to avoid over-fitting.
Our model obtains 99.6$\%$ top-1 test accuracy on 501 samples.
The training curve of the \textit{StaDNet} on the \textit{Praxis} dataset is shown in Figure~\ref{training_curve_praxis}, the normalized confusion matrix on this dataset is shown in Figure~\ref{confusion_matrix_praxis}, while the comparison of the results with the state-of-the-art is shown in Table~\ref{praxis_table}.
We also quantify the performance of our static hand gesture detector on  a test set of 4190 hand images. The overall top-1 test accuracy is found to be 98.9$\%$. The normalized confusion
matrix for 10 static hand gestures is shown in Figure~\ref{confusion_hand}.

\begin{table}[h]
	\centering
	\begin{tabular}{|c|c|}
		\hline
		\textbf{System}      & \textbf{Accuracy $\%$ (dynamic gestures)} \\ \hline
		\textit{StaDNet} (ours) & \textbf{99.60}                        \\ \hline
		Negin \textit{et al.} \cite{negin2018praxis}         & 76.61                                \\ \hline
	\end{tabular}
	\vspace{0.5em}
	\caption{Comparison of dynamic gestures recognition results on the Praxis gestures dataset; \cite{negin2018praxis} also used a similar CNN-LSTM network.}
	\vspace{-1em}
	\label{praxis_table}
\end{table}

\begin{figure*}[t!]
	\centering
	\includegraphics[width=0.99\linewidth]{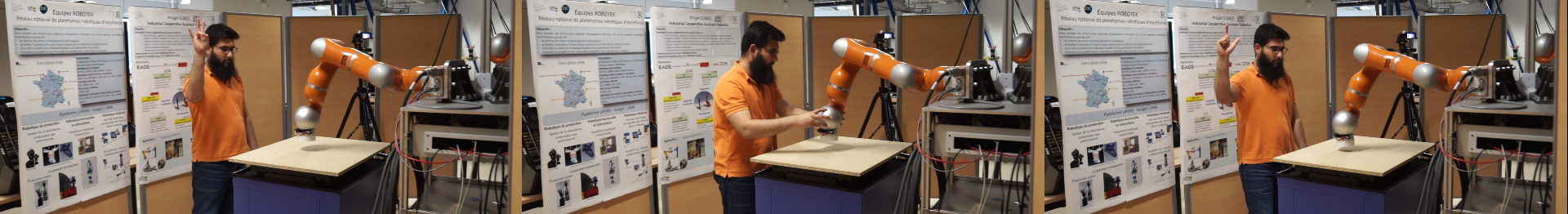}
	\caption{Snapshots of our gesture-controlled safe human-robot interaction experiment taken from \cite{mazhar2019real} with the authors' permission. The human operator manually guides the robot to waypoints in the workspace then asks the robot to \textit{record} them through a gesture. The human operator can transmit other commands to the robot like \textit{replay, stop, resume, reteach,} etc with only hand gestures.
	}
	\label{HRI_experiment}
\end{figure*}

We devised robotic experiments for gesture-controlled safe human-robot interaction tasks as already presented in \cite{mazhar2019real}. These are preliminary experiments that allow the human operator to communicate with the robot through static hand gestures in real-time while dynamic gestures integration is yet to be done.
The experiments were performed on BAZAR robot \cite{cherubini2019collaborative} which has two Kuka LWR 4+ arms with two Shadow Dexterous Hands attached at the end-effectors.
We exploited \textit{OpenPHRI} \cite{navarro2018pursuit}, which is an open-source library, to control the robot while corroborating safety of the human operator.
A finite state machine is developed to control behavior of the robot which is determined by the sensory information e.g., hand gestures, distance of the human operator from the robot, joint-torque sensing etc. The experiment is decomposed into two phases: 1) a teaching by demonstration phase, where the user manually guides the robot to a set of waypoints and 2) a replay phase, where the robot autonomously goes to every recorded waypoint to perform a given task, here force control. A video of the experiment is available online\footnote{\url{http://youtu.be/lB5vXc8LMnk}} and snapshots are given in
Figure~\ref{HRI_experiment}.

\section{Conclusion} \label{conclusion}

In this paper, a unified framework for simultaneous recognition of static hands and dynamic upper-body gestures, \textit{StaDNet} is proposed.
A novel idea of learning-based depth estimator is also presented, which predicts the distance of the person and his/her hands, exploiting only the upper-body 2D skeleton coordinates.
By virtue of this feature, monocular images are sufficient and the proposed framework does not require depth sensing.
Thus, the use of \textit{StaDNet} for gestures detection is not limited to any specialized camera and can work with most conventional RGB cameras.
Monocular images are indeed sensitive to the changing lighting conditions and might fail to work in extreme conditions e.g., during sand blasting operation in the industry or during fog and rain in the outdoors.
To develop immunity against such lighting corruptions, data augmentation strategies such as \cite{Osama2021a} can be exploited.
One might argue that employing HSV or HSL color models instead of RGB might be more appropriate to deal with changing ambient light conditions.
However, \textit{StaDNet} actually relies on \textit{OpenPose} for skeleton extraction and on the hand gesture detector from our previous work \cite{mazhar2019real}.
\textit{OpenPose} is the state-of-art in skeleton extraction from monocular camera and takes RGB images as input.
Furthermore, our static hand gesture also takes RGB images as input and performs well with 98.9\% top-1 test accuracy on 10 static hand gestures as we show in Figure \ref{confusion_hand}.
In spite of that, we are aware that HSV or HSL had been commonly used for hand segmentation in the literature by thresholding the values of Hue, Saturation and Value/Lightness.
This indeed intrigues our eagerness to train and compare the performance of deep models for hand gesture detector in this color model/space, that we plan to do in our future work.

Our pose-driven hard spatial attention mechanism directs the focus of \textit{StaDNet} on upper-body pose to model large-scale body movements of the limbs and, on the hand images for subtle hand/fingers movements.
This enables \textit{StaDNet} to out-score the existing approaches on the Chalearn 2016 dataset.
The presented weight initialization strategy addresses the imbalance in class distribution in the Chalearn 2016 dataset, thus facilitates parameters optimization for all 249 gestures.
Our static gestures detector outputs the predicted label frame-wise at approximately 21 fps with the state-of-the-art recognition accuracy.
However, class recognition for dynamic gestures is performed on isolated gestures videos, executed by an individual in the scene.
We plan to extend this work for continuous dynamic gestures recognition to demonstrate its utility in human-machine interaction.
This can be achieved in one way by developing a binary motion detector to detect start and end instances of the gestures.
Although a multi-stage training strategy is presented, we envision an end-to-end training approach for online learning of new gestures.

\acknowledgments{The research presented in this article was carried out as parts of the SOPHIA and the OpenDR projects, which have received funding from the European Union’s Horizon 2020 research and innovation programme under Grant Agreement No. 871237 and 871449 respectively.}

\end{paracol}
\reftitle{References}


\externalbibliography{yes}
\bibliography{references.bib}

\begin{thebibliography}{999}

\bibitem[Li \em{et~al.}(2018)Li, Wu, Jiang, Xu, and Liu]{li2018dynamic}
Li, G.; Wu, H.; Jiang, G.; Xu, S.; Liu, H.
\newblock Dynamic gesture recognition in the internet of things.
\newblock {\em IEEE Access} {\bf 2018}, {\em 7},~23713--23724.

\bibitem[Kofman \em{et~al.}(2005)Kofman, Wu, Luu, and
  Verma]{kofman2005teleoperation}
Kofman, J.; Wu, X.; Luu, T.J.; Verma, S.
\newblock Teleoperation of a robot manipulator using a vision-based human-robot
  interface.
\newblock {\em IEEE Trans. on Industrial Electronics} {\bf 2005}, {\em
  52},~1206--1219.

\bibitem[T{\"o}lgyessy \em{et~al.}(2017{\natexlab{a}})T{\"o}lgyessy,
  Hubinsk{\`y}, Chovanec, Ducho{\v{n}}, and Babinec]{tolgyessy2017controlling}
T{\"o}lgyessy, M.; Hubinsk{\`y}, P.; Chovanec, L.; Ducho{\v{n}}, F.; Babinec,
  A.
\newblock Controlling a group of robots to perform a common task by gestures
  only.
\newblock {\em Int J Imaging Robot} {\bf 2017}, {\em 17},~1--13.

\bibitem[T{\"o}lgyessy \em{et~al.}(2017{\natexlab{b}})T{\"o}lgyessy, Dekan,
  Ducho{\v{n}}, Rodina, Hubinsk{\`y}, and Chovanec]{tolgyessy2017foundations}
T{\"o}lgyessy, M.; Dekan, M.; Ducho{\v{n}}, F.; Rodina, J.; Hubinsk{\`y}, P.;
  Chovanec, L.
\newblock Foundations of visual linear human--robot interaction via pointing
  gesture navigation.
\newblock {\em International Journal of Social Robotics} {\bf 2017}, {\em
  9},~509--523.

\bibitem[Jung \em{et~al.}(2015)Jung, Lim, Kim, and Kong]{jung2015wearable}
Jung, P.G.; Lim, G.; Kim, S.; Kong, K.
\newblock A wearable gesture recognition device for detecting muscular
  activities based on air-pressure sensors.
\newblock {\em IEEE Trans. on Industrial Informatics} {\bf 2015}, {\em
  11},~485--494.

\bibitem[Park \em{et~al.}(2006)Park, Jung, and Kim]{park2006vision}
Park, H.S.; Jung, D.J.; Kim, H.J.
\newblock {Vision-based Game Interface using Human Gesture}.
\newblock  Pacific-Rim Symposium on Image and Video Technology. Springer,
  2006, pp. 662--671.

\bibitem[Neverova \em{et~al.}(2014)Neverova, Wolf, Taylor, and
  Nebout]{neverova2014}
Neverova, N.; Wolf, C.; Taylor, G.W.; Nebout, F.
\newblock {Multi-scale Deep Learning for Gesture Detection and Localization}.
\newblock  European Conf. on Computer Vision. Springer,  2014, pp. 474--490.

\bibitem[Gleeson \em{et~al.}(2013)Gleeson, MacLean, Haddadi, Croft, and
  Alcazar]{Gleeson2013a}
Gleeson, B.; MacLean, K.; Haddadi, A.; Croft, E.; Alcazar, J.
\newblock Gestures for industry intuitive human-robot communication from human
  observation.
\newblock  2013 8th ACM/IEEE International Conference on Human-Robot
  Interaction (HRI). IEEE,  2013, pp. 349--356.

\bibitem[Mazhar(2019)]{OpenSign_mazhar}
Mazhar, O.
\newblock {OpenSign - Kinect V2 Hand Gesture Data - American Sign Language},
  2019.

\bibitem[Starner \em{et~al.}(1998)Starner, Weaver, and Pentland]{Starner1998a}
Starner, T.; Weaver, J.; Pentland, A.
\newblock Real-time american sign language recognition using desk and wearable
  computer based video.
\newblock {\em IEEE Transactions on pattern analysis and machine intelligence}
  {\bf 1998}, {\em 20},~1371--1375.

\bibitem[Mazhar \em{et~al.}(2019)Mazhar, Navarro, Ramdani, Passama, and
  Cherubini]{mazhar2019real}
Mazhar, O.; Navarro, B.; Ramdani, S.; Passama, R.; Cherubini, A.
\newblock {A Real-time Human-Robot Interaction Framework with Robust Background
  Invariant Hand Gesture Detection}.
\newblock {\em Robotics and Computer-Integrated Manufacturing} {\bf 2019}, {\em
  60},~34--48.

\bibitem[Rensink(2000)]{rensink2000dynamic}
Rensink, R.A.
\newblock {The Dynamic Representation of Scenes}.
\newblock {\em Visual Cognition} {\bf 2000}, {\em 7},~17--42.

\bibitem[{Neto} \em{et~al.}(2013){Neto}, {Pereira}, {Pires}, and
  {Moreira}]{Neto2013a}
{Neto}, P.; {Pereira}, D.; {Pires}, J.N.; {Moreira}, A.P.
\newblock Real-time and continuous hand gesture spotting: An approach based on
  artificial neural networks.
\newblock  2013 IEEE International Conference on Robotics and Automation,
  2013, pp. 178--183.
\newblock
  doi:{\changeurlcolor{black}\href{https://doi.org/10.1109/ICRA.2013.6630573}{\detokenize{10.1109/ICRA.2013.6630573}}}.

\bibitem[{Wong} \em{et~al.}(2021){Wong}, {Juwono}, and {Khoo}]{Wong2021a}
{Wong}, W.K.; {Juwono}, F.H.; {Khoo}, B.T.T.
\newblock Multi-Features Capacitive Hand Gesture Recognition Sensor: A Machine
  Learning Approach.
\newblock {\em IEEE Sensors Journal} {\bf 2021}, {\em 21},~8441--8450.
\newblock
  doi:{\changeurlcolor{black}\href{https://doi.org/10.1109/JSEN.2021.3049273}{\detokenize{10.1109/JSEN.2021.3049273}}}.

\bibitem[Zhu and Sheng(2011)]{zhu2011motion}
Zhu, C.; Sheng, W.
\newblock Motion-and location-based online human daily activity recognition.
\newblock {\em Pervasive and Mobile Computing} {\bf 2011}, {\em 7},~256--269.

\bibitem[Laptev(2005)]{laptev2005space}
Laptev, I.
\newblock {On Space-time Interest Points}.
\newblock {\em Int. Journal of Computer Vision} {\bf 2005}, {\em 64},~107--123.

\bibitem[Doll{\'a}r \em{et~al.}(2005)Doll{\'a}r, Rabaud, Cottrell, and
  Belongie]{dollar2005behavior}
Doll{\'a}r, P.; Rabaud, V.; Cottrell, G.; Belongie, S.
\newblock {Behavior Recognition via Sparse Spatio-Temporal Features}.
\newblock  2005 IEEE Int. Workshop on Visual Surveillance and Performance
  Evaluation of Tracking and Surveillance,  2005, pp. 65--72.

\bibitem[Wang \em{et~al.}()Wang, Ullah, Klaser, Laptev, and Schmid]{Wang2009}
Wang, H.; Ullah, M.M.; Klaser, A.; Laptev, I.; Schmid, C.
\newblock {Evaluation of local spatio-temporal features for action
  recognition}.
\newblock {\em Proc. of the British Machine Vision Conf. 2009}, pp.
  124.1--124.11.

\bibitem[{Wang} \em{et~al.}(2011){Wang}, {Kläser}, {Schmid}, and
  {Liu}]{Wang2011}
{Wang}, H.; {Kläser}, A.; {Schmid}, C.; {Liu}, C.
\newblock {Action Recognition by Dense Trajectories}.
\newblock  IEEE Conf. on Computer Vision and Pattern Recognition,  2011, pp.
  3169--3176.

\bibitem[{Wang} and {Schmid}(2013)]{Wang2013}
{Wang}, H.; {Schmid}, C.
\newblock {Action Recognition with Improved Trajectories}.
\newblock  2013 IEEE Int. Conf. on Computer Vision,  2013, pp. 3551--3558.

\bibitem[Wang \em{et~al.}(2016)Wang, Oneata, Verbeek, and
  Schmid]{wang2016robust}
Wang, H.; Oneata, D.; Verbeek, J.; Schmid, C.
\newblock {A Robust and Efficient Video Representation for Action Recognition}.
\newblock {\em Int. Journal of Computer Vision} {\bf 2016}, {\em
  119},~219--238.

\bibitem[S{\'a}nchez \em{et~al.}(2013)S{\'a}nchez, Perronnin, Mensink, and
  Verbeek]{sanchez2013image}
S{\'a}nchez, J.; Perronnin, F.; Mensink, T.; Verbeek, J.
\newblock {Image classification with the Fisher Vector: Theory and Practice}.
\newblock {\em Int. Journal of Computer Vision} {\bf 2013}, {\em
  105},~222--245.

\bibitem[Kantorov and Laptev(2014)]{kantorov2014efficient}
Kantorov, V.; Laptev, I.
\newblock {Efficient Feature Extraction, Encoding and Classification for Action
  Recognition}.
\newblock  Proc. of the IEEE Conf. on Computer Vision and Pattern Recognition,
  2014, pp. 2593--2600.

\bibitem[Wu \em{et~al.}(2012)Wu, Zhu, and Shao]{wu2012one}
Wu, D.; Zhu, F.; Shao, L.
\newblock One shot learning gesture recognition from rgbd images.
\newblock  2012 IEEE Computer Society Conference on Computer Vision and Pattern
  Recognition Workshops. IEEE,  2012, pp. 7--12.

\bibitem[Otsu(1979)]{otsu1979threshold}
Otsu, N.
\newblock A threshold selection method from gray-level histograms.
\newblock {\em IEEE transactions on systems, man, and cybernetics} {\bf 1979},
  {\em 9},~62--66.

\bibitem[Fanello \em{et~al.}(2013)Fanello, Gori, Metta, and
  Odone]{fanello2013keep}
Fanello, S.R.; Gori, I.; Metta, G.; Odone, F.
\newblock Keep It Simple And Sparse: Real-Time Action Recognition.
\newblock {\em Journal of Machine Learning Research} {\bf 2013}, {\em 14}.

\bibitem[Kone{\v{c}}n{\`y} and Hagara(2014)]{konevcny2014one}
Kone{\v{c}}n{\`y}, J.; Hagara, M.
\newblock One-shot-learning gesture recognition using hog-hof features.
\newblock {\em The Journal of Machine Learning Research} {\bf 2014}, {\em
  15},~2513--2532.

\bibitem[Wan \em{et~al.}(2013)Wan, Ruan, Li, and Deng]{wan2013one}
Wan, J.; Ruan, Q.; Li, W.; Deng, S.
\newblock One-shot learning gesture recognition from RGB-D data using bag of
  features.
\newblock {\em The Journal of Machine Learning Research} {\bf 2013}, {\em
  14},~2549--2582.

\bibitem[He \em{et~al.}(2016)He, Zhang, Ren, and Sun]{he2016deep}
He, K.; Zhang, X.; Ren, S.; Sun, J.
\newblock {Deep Residual Learning for Image Recognition}.
\newblock  Proc. of the IEEE Conf. on Computer Vision and Pattern Recognition,
  2016, pp. 770--778.

\bibitem[Simonyan and Zisserman(2014)]{simonyan2014very}
Simonyan, K.; Zisserman, A.
\newblock {Very Deep Convolutional Networks for Large-scale Image Recognition}.
\newblock {\em arXiv preprint arXiv:1409.1556} {\bf 2014}.

\bibitem[Ji \em{et~al.}(2012)Ji, Xu, Yang, and Yu]{ji20123d}
Ji, S.; Xu, W.; Yang, M.; Yu, K.
\newblock {3D Convolutional Neural Networks for Human Action Recognition}.
\newblock {\em IEEE Trans. on Pattern Analysis and Machine Intelligence} {\bf
  2012}, {\em 35},~221--231.

\bibitem[Baccouche \em{et~al.}(2011)Baccouche, Mamalet, Wolf, Garcia, and
  Baskurt]{baccouche2011sequential}
Baccouche, M.; Mamalet, F.; Wolf, C.; Garcia, C.; Baskurt, A.
\newblock {Sequential Deep Learning for Human Action Recognition}.
\newblock  Int. Workshop on Human Behavior Understanding. Springer,  2011, pp.
  29--39.

\bibitem[Hochreiter and Schmidhuber(1997)]{hochreiter1997long}
Hochreiter, S.; Schmidhuber, J.
\newblock {Long Short-Term Memory}.
\newblock {\em Neural Computation} {\bf 1997}, {\em 9},~1735--1780.

\bibitem[Karpathy \em{et~al.}()Karpathy, Toderici, Shetty, Leung, Sukthankar,
  and Li]{Karpathy2014}
Karpathy, A.; Toderici, G.; Shetty, S.; Leung, T.; Sukthankar, R.; Li, F.F.
\newblock {Large-scale Video Classification with Convolutional Neural
  Networks}.
\newblock {\em Proc. of the IEEE Computer Society Conf. on Computer Vision and
  Pattern Recognition}, pp. 1725--1732.

\bibitem[Xu \em{et~al.}(2018)Xu, Kang, Sun, Feng, Saenko, and
  Darrell]{xu2018similarity}
Xu, H.; Kang, B.; Sun, X.; Feng, J.; Saenko, K.; Darrell, T.
\newblock Similarity r-c3d for few-shot temporal activity detection.
\newblock {\em arXiv preprint arXiv:1812.10000} {\bf 2018}.

\bibitem[Zheng \em{et~al.}(2018)Zheng, Cao, Zhang, Zhen, and Su]{zheng2018deep}
Zheng, J.; Cao, X.; Zhang, B.; Zhen, X.; Su, X.
\newblock Deep ensemble machine for video classification.
\newblock {\em IEEE Trans. on Neural Networks and Learning Systems} {\bf 2018},
  {\em 30},~553--565.

\bibitem[Neverova \em{et~al.}(2013)Neverova, Wolf, Paci, Sommavilla, Taylor,
  and Nebout]{neverova2013multi}
Neverova, N.; Wolf, C.; Paci, G.; Sommavilla, G.; Taylor, G.; Nebout, F.
\newblock {A Multi-scale Approach to Gesture Detection and Recognition}.
\newblock  Proc. of the IEEE Int. Conf. on Computer Vision Workshops,  2013,
  pp. 484--491.

\bibitem[Miao \em{et~al.}(2017)Miao, Li, Ouyang, Ma, Xu, Shi, and
  Cao]{miao2017multimodal}
Miao, Q.; Li, Y.; Ouyang, W.; Ma, Z.; Xu, X.; Shi, W.; Cao, X.
\newblock {Multimodal Gesture Recognition based on the ResC3D Network}.
\newblock  Proc. of the IEEE Int. Conf. on Computer Vision,  2017, pp.
  3047--3055.

\bibitem[Wan \em{et~al.}(2016)Wan, Zhao, Zhou, Guyon, Escalera, and
  Li]{wan2016chalearn}
Wan, J.; Zhao, Y.; Zhou, S.; Guyon, I.; Escalera, S.; Li, S.Z.
\newblock {Chalearn Looking at People RGB-D Isolated and Continuous Datasets
  for Gesture Recognition}.
\newblock  Proc. of the IEEE Conf. on Computer Vision and Pattern Recognition
  Workshops,  2016, pp. 56--64.

\bibitem[Tran \em{et~al.}(2017)Tran, Ray, Shou, Chang, and
  Paluri]{tran2017convnet}
Tran, D.; Ray, J.; Shou, Z.; Chang, S.F.; Paluri, M.
\newblock {ConvNet Architecture Search for Spatiotemporal Feature Learning}.
\newblock {\em arXiv preprint arXiv:1708.05038} {\bf 2017}.

\bibitem[Simonyan and Zisserman(2014)]{simonyan2014two}
Simonyan, K.; Zisserman, A.
\newblock {Two-stream Convolutional Networks for Action Recognition in Videos}.
\newblock  Advances in Neural Information Processing Systems,  2014, pp.
  568--576.

\bibitem[Goodale and Milner(1992)]{goodale1992separate}
Goodale, M.A.; Milner, A.D.
\newblock {Separate Visual Pathways for Perception and Action}.
\newblock {\em Trends in Neurosciences} {\bf 1992}, {\em 15},~20--25.

\bibitem[Wang \em{et~al.}(2015)Wang, Qiao, and Tang]{wang2015action}
Wang, L.; Qiao, Y.; Tang, X.
\newblock {Action Recognition with Trajectory-Pooled Deep-Convolutional
  Descriptors}.
\newblock  Proc. of the IEEE Conf. on Computer Vision and Pattern Recognition,
  2015, pp. 4305--4314.

\bibitem[Yue-Hei~Ng \em{et~al.}(2015)Yue-Hei~Ng, Hausknecht, Vijayanarasimhan,
  Vinyals, Monga, and Toderici]{yue2015beyond}
Yue-Hei~Ng, J.; Hausknecht, M.; Vijayanarasimhan, S.; Vinyals, O.; Monga, R.;
  Toderici, G.
\newblock {Beyond Short Snippets: Deep Networks for Video Classification}.
\newblock  Proc. of the IEEE Conf. on Computer Vision and Pattern Recognition,
  2015, pp. 4694--4702.

\bibitem[Donahue \em{et~al.}(2015)Donahue, Anne~Hendricks, Guadarrama,
  Rohrbach, Venugopalan, Saenko, and Darrell]{donahue2015long}
Donahue, J.; Anne~Hendricks, L.; Guadarrama, S.; Rohrbach, M.; Venugopalan, S.;
  Saenko, K.; Darrell, T.
\newblock {Long-Term Recurrent Convolutional Networks for Visual Recognition
  and Description}.
\newblock  Proc. of the IEEE Conf. on Computer Vision and Pattern Recognition,
  2015, pp. 2625--2634.

\bibitem[Xingjian \em{et~al.}(2015)Xingjian, Chen, Wang, Yeung, Wong, and
  Woo]{xingjian2015convolutional}
Xingjian, S.; Chen, Z.; Wang, H.; Yeung, D.Y.; Wong, W.K.; Woo, W.c.
\newblock {Convolutional LSTM network: A Machine Learning Approach for
  Precipitation Nowcasting}.
\newblock  Advances in Neural Information Processing Systems,  2015, pp.
  802--810.

\bibitem[Zhu \em{et~al.}(2019)Zhu, Zhang, Yang, Mei, Shah, Bennamoun, and
  Shen]{zhu2019redundancy}
Zhu, G.; Zhang, L.; Yang, L.; Mei, L.; Shah, S.A.A.; Bennamoun, M.; Shen, P.
\newblock {Redundancy and Attention in Convolutional LSTM for Gesture
  Recognition}.
\newblock {\em IEEE Trans. on Neural Networks and Learning Systems} {\bf 2019}.

\bibitem[Yeung \em{et~al.}(2018)Yeung, Russakovsky, Jin, Andriluka, Mori, and
  Fei-Fei]{yeung2018every}
Yeung, S.; Russakovsky, O.; Jin, N.; Andriluka, M.; Mori, G.; Fei-Fei, L.
\newblock {Every Moment Counts: Dense Detailed Labeling of Actions in Complex
  Videos}.
\newblock {\em Int. Journal of Computer Vision} {\bf 2018}, {\em
  126},~375--389.

\bibitem[Krizhevsky \em{et~al.}(2012)Krizhevsky, Sutskever, and
  Hinton]{krizhevsky2012imagenet}
Krizhevsky, A.; Sutskever, I.; Hinton, G.E.
\newblock {ImageNet Classification with Deep Convolutional Neural Networks}.
\newblock  Advances in Neural Information Processing Systems,  2012, pp.
  1097--1105.

\bibitem[Idrees \em{et~al.}(2017)Idrees, Zamir, Jiang, Gorban, Laptev,
  Sukthankar, and Shah]{idrees2017thumos}
Idrees, H.; Zamir, A.R.; Jiang, Y.G.; Gorban, A.; Laptev, I.; Sukthankar, R.;
  Shah, M.
\newblock {The THUMOS Challenge on Action Recognition for Videos “In the
  Wild”}.
\newblock {\em Computer Vision and Image Understanding} {\bf 2017}, {\em
  155},~1--23.

\bibitem[Mnih \em{et~al.}(2014)Mnih, Heess, Graves, et~al.]{mnih2014recurrent}
Mnih, V.; Heess, N.; Graves, A.; others.
\newblock {Recurrent Models of Visual Attention}.
\newblock  Advances in Neural Information Processing Systems,  2014, pp.
  2204--2212.

\bibitem[Baradel \em{et~al.}(2017{\natexlab{a}})Baradel, Wolf, and
  Mille]{baradel2017pose}
Baradel, F.; Wolf, C.; Mille, J.
\newblock {Pose-conditioned Spatio-temporal Attention for Human Action
  Recognition}.
\newblock {\em arXiv preprint arXiv:1703.10106} {\bf 2017}.

\bibitem[Baradel \em{et~al.}(2017{\natexlab{b}})Baradel, Wolf, and
  Mille]{baradel2017human}
Baradel, F.; Wolf, C.; Mille, J.
\newblock {Human Action Recognition: Pose-based Attention Draws Focus to
  Hands}.
\newblock  Proc. of the IEEE Int. Conf. on Computer Vision,  2017, pp.
  604--613.

\bibitem[Zheng \em{et~al.}(2020)Zheng, An, Wu, and Ruan]{zheng2020global}
Zheng, Z.; An, G.; Wu, D.; Ruan, Q.
\newblock Global and Local Knowledge-Aware Attention Network for Action
  Recognition.
\newblock {\em IEEE Trans. on Neural Networks and Learning Systems} {\bf 2020}.

\bibitem[Narayana \em{et~al.}(2018)Narayana, Beveridge, and
  Draper]{narayana2018gesture}
Narayana, P.; Beveridge, R.; Draper, B.A.
\newblock {Gesture Recognition: Focus on the Hands}.
\newblock  Proc. of the IEEE Conf. on Computer Vision and Pattern Recognition,
  2018, pp. 5235--5244.

\bibitem[Negin \em{et~al.}(2018)Negin, Rodriguez, Koperski, Kerboua,
  Gonz{\`a}lez, Bourgeois, Chapoulie, Robert, and Bremond]{negin2018praxis}
Negin, F.; Rodriguez, P.; Koperski, M.; Kerboua, A.; Gonz{\`a}lez, J.;
  Bourgeois, J.; Chapoulie, E.; Robert, P.; Bremond, F.
\newblock {PRAXIS: Towards Automatic Cognitive Assessment Using Gesture
  Recognition}.
\newblock {\em Expert Systems with Applications} {\bf 2018}.

\bibitem[Cao \em{et~al.}(2017)Cao, Simon, Wei, and Sheikh]{cao2017realtime}
Cao, Z.; Simon, T.; Wei, S.E.; Sheikh, Y.
\newblock {Realtime Multi-Person 2D Pose Estimation using Part Affinity
  Fields}.
\newblock  IEEE Conf. on Computer Vision and Pattern Recognition,  2017.

\bibitem[Panteleris \em{et~al.}(2018)Panteleris, Oikonomidis, and
  Argyros]{panteleris2018using}
Panteleris, P.; Oikonomidis, I.; Argyros, A.
\newblock {Using a Single RGB Frame for Real time 3D Hand Pose Estimation in
  the Wild}.
\newblock  2018 IEEE Winter Conf. on Applications of Computer Vision (WACV).
  IEEE,  2018, pp. 436--445.

\bibitem[Wang \em{et~al.}(2017)Wang, Wang, Song, and Li]{wang2017large}
Wang, H.; Wang, P.; Song, Z.; Li, W.
\newblock {Large-scale Multimodal Gesture Recognition using Heterogeneous
  Networks}.
\newblock  Proc. of the IEEE Int. Conf. on Computer Vision,  2017, pp.
  3129--3137.

\bibitem[Wan \em{et~al.}(2017)Wan, Escalera, Anbarjafari, Jair~Escalante,
  Bar{\'o}, Guyon, Madadi, Allik, Gorbova, Lin, et~al.]{wan2017results}
Wan, J.; Escalera, S.; Anbarjafari, G.; Jair~Escalante, H.; Bar{\'o}, X.;
  Guyon, I.; Madadi, M.; Allik, J.; Gorbova, J.; Lin, C.; others.
\newblock {Results and Analysis of Chalearn LAP Multi-modal Isolated and
  Continuous Gesture Recognition, and Real versus Fake Expressed Emotions
  Challenges}.
\newblock  Proc. of the IEEE Int. Conf. on Computer Vision,  2017, pp.
  3189--3197.

\bibitem[Hu \em{et~al.}(2018)Hu, Lin, and Hsiu]{hu2018learning}
Hu, T.K.; Lin, Y.Y.; Hsiu, P.C.
\newblock {Learning Adaptive Hidden Layers for Mobile Gesture Recognition}.
\newblock  Thirty-Second AAAI Conf. on Artificial Intelligence,  2018.

\bibitem[Hendrycks \em{et~al.}(2020)Hendrycks, Basart, Mu, Kadavath, Wang,
  Dorundo, Desai, Zhu, Parajuli, Guo, et~al.]{Hendrycks2020a}
Hendrycks, D.; Basart, S.; Mu, N.; Kadavath, S.; Wang, F.; Dorundo, E.; Desai,
  R.; Zhu, T.; Parajuli, S.; Guo, M.; others.
\newblock The many faces of robustness: A critical analysis of
  out-of-distribution generalization.
\newblock {\em arXiv preprint arXiv:2006.16241} {\bf 2020}.

\bibitem[Cherubini \em{et~al.}(2019)Cherubini, Passama, Navarro, Sorour,
  Khelloufi, Mazhar, Tarbouriech, Zhu, Tempier, Crosnier,
  et~al.]{cherubini2019collaborative}
Cherubini, A.; Passama, R.; Navarro, B.; Sorour, M.; Khelloufi, A.; Mazhar, O.;
  Tarbouriech, S.; Zhu, J.; Tempier, O.; Crosnier, A.; others.
\newblock A collaborative robot for the factory of the future: Bazar.
\newblock {\em The International Journal of Advanced Manufacturing Technology}
  {\bf 2019}, {\em 105},~3643--3659.

\bibitem[Navarro \em{et~al.}(2018)Navarro, Fonte, Fraisse, Poisson, and
  Cherubini]{navarro2018pursuit}
Navarro, B.; Fonte, A.; Fraisse, P.; Poisson, G.; Cherubini, A.
\newblock In pursuit of safety: An open-source library for physical human-robot
  interaction.
\newblock {\em IEEE Robotics \& Automation Magazine} {\bf 2018}, {\em
  25},~39--50.

\bibitem[Mazhar and Kober(2021)]{Osama2021a}
Mazhar, O.; Kober, J.
\newblock Random Shadows and Highlights: A new data augmentation method for
  extreme lighting conditions.
\newblock {\em arXiv preprint arXiv:2101.05361} {\bf 2021}.

\end{thebibliography}

\end{document}